\documentclass[10pt,twocolumn,letterpaper]{article}

\usepackage{cvpr}              
\usepackage[accsupp]{axessibility}
\usepackage{graphicx}

\usepackage{amsmath}
\usepackage{amssymb}
\usepackage{booktabs}
\usepackage{multirow}
\usepackage{algorithm}
\usepackage{algorithmic}

\newcommand*{\affaddr}[1]{#1} 
\newcommand*{\affmark}[1][*]{\textsuperscript{#1}}
\renewcommand\thefootnote{}

\usepackage{enumitem}
\newenvironment{tight_itemize}{
\begin{itemize}[leftmargin=20pt]
\setlength{\topsep}{0pt}
\setlength{\itemsep}{0pt}
\setlength{\parskip}{0pt}
\setlength{\parsep}{0pt}
}{\end{itemize}}

\usepackage[dvipsnames]{xcolor}
\usepackage[pagebackref,breaklinks,colorlinks]{hyperref}

\usepackage[capitalize]{cleveref}
\crefname{section}{Sec.}{Secs.}
\Crefname{section}{Section}{Sections}
\Crefname{table}{Table}{Tables}
\crefname{table}{Tab.}{Tabs.}

\def\cvprPaperID{6529} 
\def\confName{CVPR}
\def\confYear{2022}

\begin{document}

\title{Killing Two Birds with One Stone: \\
Efficient and Robust Training of Face Recognition CNNs by Partial FC}

\author{
Xiang An \affmark[1,3] \qquad Jiankang Deng \textsuperscript{*} \affmark[2,3] \qquad Jia Guo \affmark[3] \\ \qquad Ziyong Feng \affmark[1] \qquad XuHan Zhu \affmark[4] \qquad Jing Yang\affmark[3]  \qquad Tongliang Liu\affmark[5]\\
\affaddr{\affmark[1]DeepGlint} \qquad
\affaddr{\affmark[2]Huawei} \qquad
\affaddr{\affmark[3]InsightFace} \qquad \\
\affaddr{\affmark[4]Peng Cheng Laboratory}\qquad 
\affaddr{\affmark[5]University of Sydney}\\
{\tt\small \{xiangan,ziyongfeng\}@deepglint.com,  tongliang.liu@sydney.edu.au} \\
{\tt\small \{jiankangdeng,guojia,zhuxuhan.research,y.jing2016\}@gmail.com}
}

\maketitle

\begin{abstract}
Learning discriminative deep feature embeddings by using million-scale in-the-wild datasets and margin-based softmax loss is the current state-of-the-art approach for face recognition. However, the memory and computing cost of the Fully Connected (FC) layer linearly scales up to the number of identities in the training set. Besides, the large-scale training data inevitably suffers from inter-class conflict and long-tailed distribution. In this paper, we propose a sparsely updating variant of the FC layer, named Partial FC (PFC). In each iteration, positive class centers and a random subset of negative class centers are selected to compute the margin-based softmax loss. All class centers are still maintained throughout the whole training process, but only a subset is selected and updated in each iteration. Therefore, the computing requirement, the probability of inter-class conflict, and the frequency of passive update on tail class centers, are dramatically reduced. Extensive experiments 
across different training data and backbones (\eg CNN and ViT) confirm the effectiveness, robustness and efficiency of the proposed PFC. 
The source code is available at \url{https://github.com/deepinsight/insightface/tree/master/recognition}.
\footnote{\textsuperscript{*} corresponding author. InsightFace is a nonprofit Github project for 2D and 3D face analysis.} 
\setcounter{footnote}{0}
\renewcommand\thefootnote{\arabic{footnote}}
\end{abstract}

\section{Introduction}

Face recognition is playing an increasingly important role in modern life and has been widely used in many real-world applications, such as face authentication on mobile devices.
Recently, face recognition has witnessed great advance along with the collection of large-scale training datasets \cite{cao2018vggface2,zhu2021webface260m}, the evolution of network architectures \cite{schroff2015facenet,he2016deep}, and the design of margin-based and mining-based loss functions \cite{schroff2015facenet,liu2017sphereface,tencent2018CosineFace,wang2018additive,deng2019arcface,wang2019mis,huang2020curricularface,xu2021consistent}.

\begin{figure}
\centering
\includegraphics[width=0.24\textwidth]{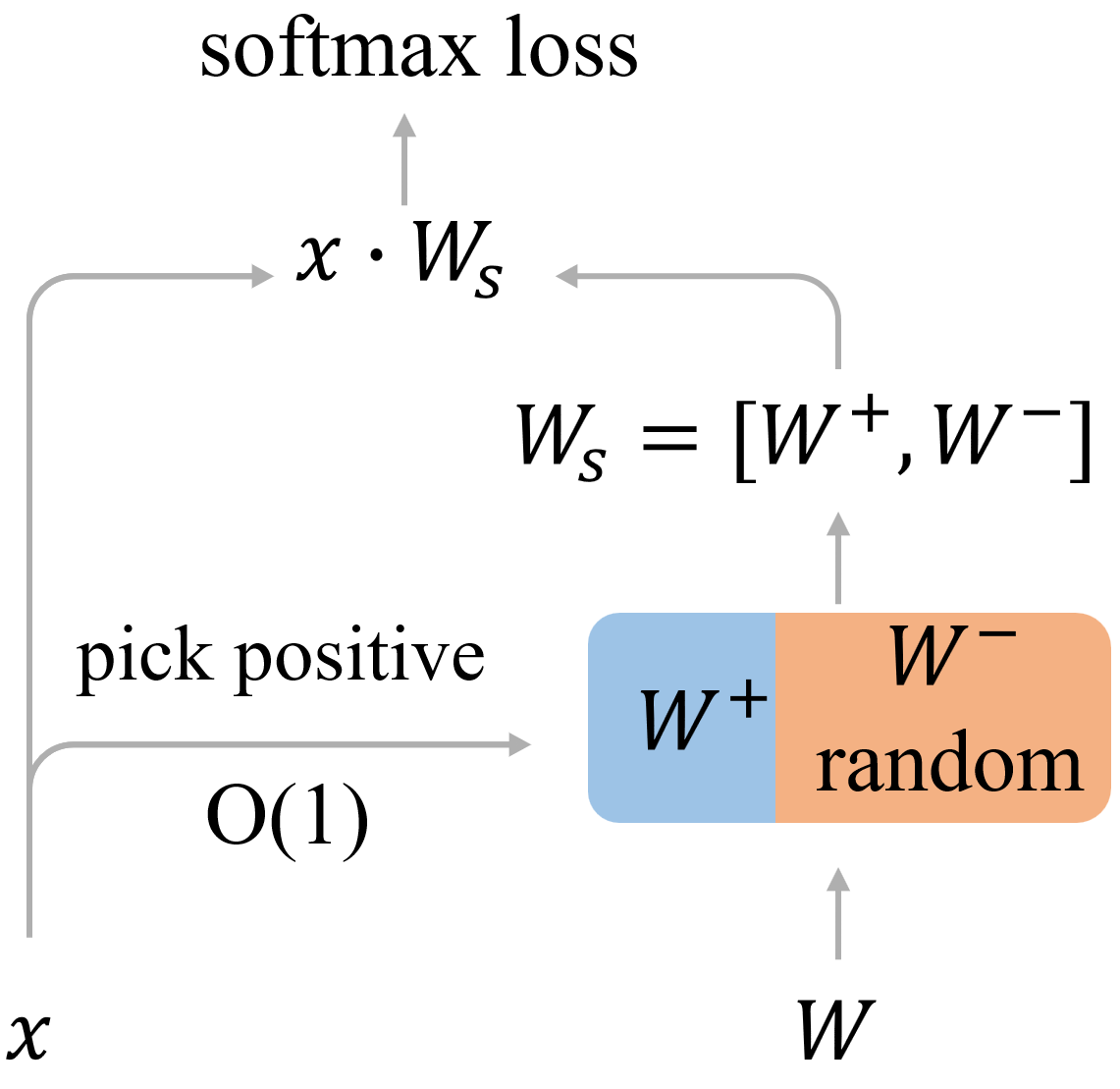}
\vspace{-2mm}
\caption{PFC picks the positive center by using the label and randomly selects a significantly reduced number of negative centers to calculate partial image-to-class similarities. PFC kills two birds (efficiency and robustness) with one stone (partial sampling). }
\vspace{-4mm}
\label{fig:simpleidea}
\end{figure}

Even though the softmax loss \cite{cao2018vggface2} and its margin-based \cite{liu2017sphereface,tencent2018CosineFace,wang2018additive,deng2019arcface} or mining-based \cite{wang2019mis,huang2020curricularface,xu2021consistent} variants achieved state-of-the-art performance on deep face recognition, the training difficulty accumulates along with the growth of identities in the training data, as the memory and computing consumption of the Fully Connected (FC) layer linearly scales up to the number of identities in the training set. When there are large-scale identities in the training dataset, the cost of storage and calculation of the final linear matrix can easily exceed the capabilities of current GPUs, resulting in tremendous training time or even a training failure.

To break the computing resource constraint, the most straightforward solution is to reduce the number of classes used during training. Zhang \etal \cite{zhang2018accelerated} propose to use a hashing forest to partition the space of class weights into small cells but the complexity of walking through the forest to find the closest cell is O(logN). Li \etal \cite{li2021virtual} randomly split training identities into groups and identities from each group share one anchor, which is used to construct the virtual fully-connected layer. Even though Virtual FC reduces the FC parameters by more than 100 times, there is an obvious performance drop compared to the conventional FC solution. SST \cite{du2020semi} and DCQ \cite{bi2021dcq} directly abandon the FC layer and employ a momentum-updated network to produce class weights. However, the negative class number is constrained to past several hundred steps and two networks need to be maintained in the GPU.

Besides the training difficulty on large-scale datasets, celebrity images gathered from the internet and cleaned by automatic methods \cite{zhu2021webface260m,MillionCelebs} exhibit long-tailed distribution \cite{liu2019adaptiveface,zhong2019unequal} as well as label noise \cite{wang2018devil}. Some well-known celebrities have abundant images (head classes) from the search engine while most celebrities have only a few images (tail classes) on the web. To keep hard training samples, the thresholds used in intra-class and inter-class cleaning steps in \cite{zhu2021webface260m} are relatively relaxed, leaving label flip noises in the WebFace42M dataset. Wang \etal \cite{wang2018devil} point out that label flips deteriorate the model's performance heavier than outliers as the margin-based softmax loss can not easily handle inter-class conflict during training.

To alleviate the above-motioned problems, we propose a sparsely updated fully connected layer, named Partial FC (PFC), for training large-scale face recognition. In the proposed PFC, the conventional FC layer is still maintained throughout the whole training process but the updating frequency is significantly decreased as we only sample parts of negative class centers in each iteration. As illustrated in Fig.~\ref{fig:simpleidea}, positive class centers are selected and a subset of negative class centers are randomly selected to compute the margin-based softmax loss. As only a subset of inter classes is selected for each iteration, the computing requirement, the frequency of passive update on tail class centers, and the probability of inter-class conflict are dramatically reduced. Extensive experiments across different training datasets and backbones (\eg CNN \cite{he2016deep} and ViT \cite{dosovitskiy2020image}) confirm the effectiveness, robustness and efficiency of the proposed PFC under a large range of sampling ratios.
The advantages of the proposed PFC can be summarized as follows:
\vspace{-0.2cm}
\begin{tight_itemize}
\item {\bf Efficient.} Under the high-performance mode, PFC-0.1 (sampling ratio) applied to ResNet100 can efficiently train 10M identities on a single server with around 2.5K samples per second, which is five times faster than the model parallel solution. Under the ultra-fast mode, the sampling ratio of PFC can be decreased to an extremely low status (around $0.01$) where no extra negative class is selected. For PFC-0.008 with ResNet100 trained on WebFace42M, the computation cost on the FC layer can be almost neglected while the verification accuracy on IJB-C reaches $97.51\%$. 
\item {\bf Robust.} PFC is amazingly robust under inter-class conflict, label-flip noise, and real-world long-tailed distribution. Assisted by a simple online abnormal inter-class filtering, PFC can further improve robustness under heavy inter-class conflicts.
\item {\bf Accurate.} The proposed PFC has obtained state-of-the-art performance on different benchmarks,
achieving $98.00\%$ on IJB-C and $97.85\%$ on MFR-all. 
\end{tight_itemize}

\section{Related Work}
\noindent{\bf Margin-based Deep Face Recognition.}
The pioneering margin-based face recognition network \cite{schroff2015facenet} uses the triplet loss in the Euclidean space. However, the training procedure is very challenging due to the combinatorial explosion in the number of triplets. By contrast, margin-based softmax methods~\cite{liu2017sphereface,tencent2018CosineFace,wang2018additive,deng2019arcface} focus on incorporating margin penalty into a more feasible framework, softmax loss, and achieve impressive performance. To further improve the margin-based softmax loss, recent works focus on the exploration of adaptive parameters \cite{zhang2019p2sgrad,zhang2019adacos,liu2019adaptiveface}, inter-class regularization \cite{zhao2019regularface,duan2019uniformface}, mining \cite{wang2019mis,huang2020curricularface,xu2021consistent}, grouping \cite{kim2020groupface}, etc.
To accelerate margin-based softmax loss, a virtual fully-connected layer is proposed by \cite{li2021virtual} to reduce the FC parameters by more than 100 times. In addition, DCQ \cite{bi2021dcq} directly abandons the FC layer and employs a momentum-updated network to produce class weights. 

\noindent{\bf Robust Face Recognition Training under Noise and Long-tail Distribution.}
Most of the face recognition datasets~\cite{cao2018vggface2,zhu2021webface260m} are downloaded from the Internet by searching a pre-defined celebrity list, and the original labels are likely to be ambiguous and inaccurate~\cite{wang2018devil}. As accurate manual annotations are expensive~\cite{wang2018devil}, learning with massive noisy data has recently drawn much attention in face recognition~\cite{wu2018light,hu2019noise,zhong2019unequal,wang2019co,deng2020sub,zhang2021adaptive}. To improve robustness under noises, recent methods attempt to design noise-tolerant loss functions (\eg computing time-varying weights for samples~\cite{hu2019noise}, designing piece-wise loss functions~\cite{zhong2019unequal} according to model's predictions, and relaxing the constraint of intra-class compactness~\cite{deng2020sub}), explore consistent predictions from twin networks~\cite{wang2019co}, and employ meta-supervision for adaptive label noise cleaning~\cite{zhang2021adaptive}. Besides the label noise, web data are usually long-tail distributed. To alleviate long-tailed distribution, recent methods attempt to either improve the margin values for the tail classes \cite{liu2019adaptiveface} or recall the benefit from sample-to-sample comparisons \cite{zhu2019largeijcv,du2020semi,bi2021dcq,deng2021variational}.

\section{Methodology}
This section starts with the limitation analysis of the conventional FC Layer. Then, these limitations motivate the proposal of a more efficient and robust training method, called Partial FC (PFC). Through learning dynamic analysis on both clean and noisy training data, we finally achieve a deeper understanding of the role of inter-class interaction.

\subsection{Revisiting FC Layer}
In this sub-section, we first discuss the optimization procedure of the FC layer. Then, we discuss three drawbacks of the FC layer based on the gradient analysis.

The most widely used classification loss function for face recognition, \ie, softmax loss, is presented as follow:
\begin{equation}
\mathcal{L}=-\frac{1}{B}{\sum\limits_{i=1}^{B}}\log\frac{e^{W^T_{y_i} x_i}}{e^{W^T_{y_i} x_i}+\sum_{j=1,j\neq  y_i}^{C}e^{W^T_j x_i}},
\label{eq:softmax}
\end{equation}
where $W_j\in\mathbb{R}^D$ denotes the $j$-th column of class-wise centers, $x_i\in\mathbb{R}^D$ denotes the feature of the $i$-th sample belonging to the $y_i$-th class, $D$ is the feature dimension, $C$ is the class number, and $B$ is the batch size. 

From the view of features, the network will be updated towards a direction that the feature will be close to the ground-truth center and far apart from all other centers. To illustrate the feature's gradient in a more straightforward way, we denote the probability and center of the ground truth as $p^+$ and $W^+$ while other negative probabilities and centers as $p^-_j$ and $W^-_j$:
\begin{gather}
\frac{\partial \mathcal{L}}{\partial {x}_{i}} = -((1-p^+)W^+ - {\sum_{j=1 ,j\ne y_i}^{C}}   p^-_j W^-_j).  
\label{x-gradient} 
\end{gather}

From the view of centers, the center $W_j$ belonging to $j$-th class will be updated towards a direction that is close to sample features of $j$-th class and far apart from sample features of other classes:
\begin{gather}
W^t_j = W^{t-1}_j + \eta(\sum_{i\in \mathbb{B} ^+} (1-p_i^+)x_i^+ - \sum_{i\in \mathbb{B}^-}  p^-_i x^-_i), 
\label{PL-update} 
\end{gather}
where $\eta$ is the learning rate, $t$ is the iteration number, $\mathbb{B}^+$ represents all samples belonging to $j$-th class, $\mathbb{B}^-$ stands for all samples of other classes, and $\left | \mathbb{B}^+ \right | + \left | \mathbb{B}^- \right |$ equals to the batch size $B$. 

\begin{figure}
\centering
\subfloat[Inter-class Conflict]{
\label{fig:webfaceinterclassconflict}
\includegraphics[height=0.16\textwidth]{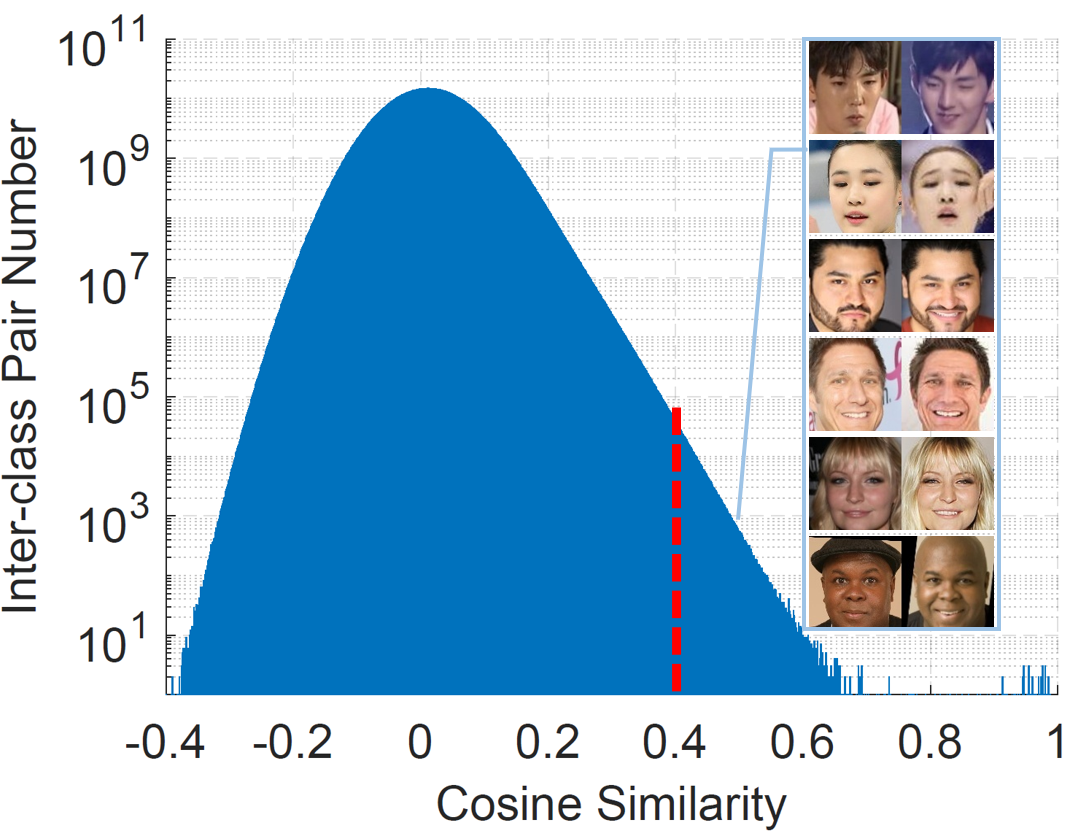}}
\subfloat[Long-tail Distribution]{
\label{fig:webfacelongtail}
\includegraphics[height=0.16\textwidth]{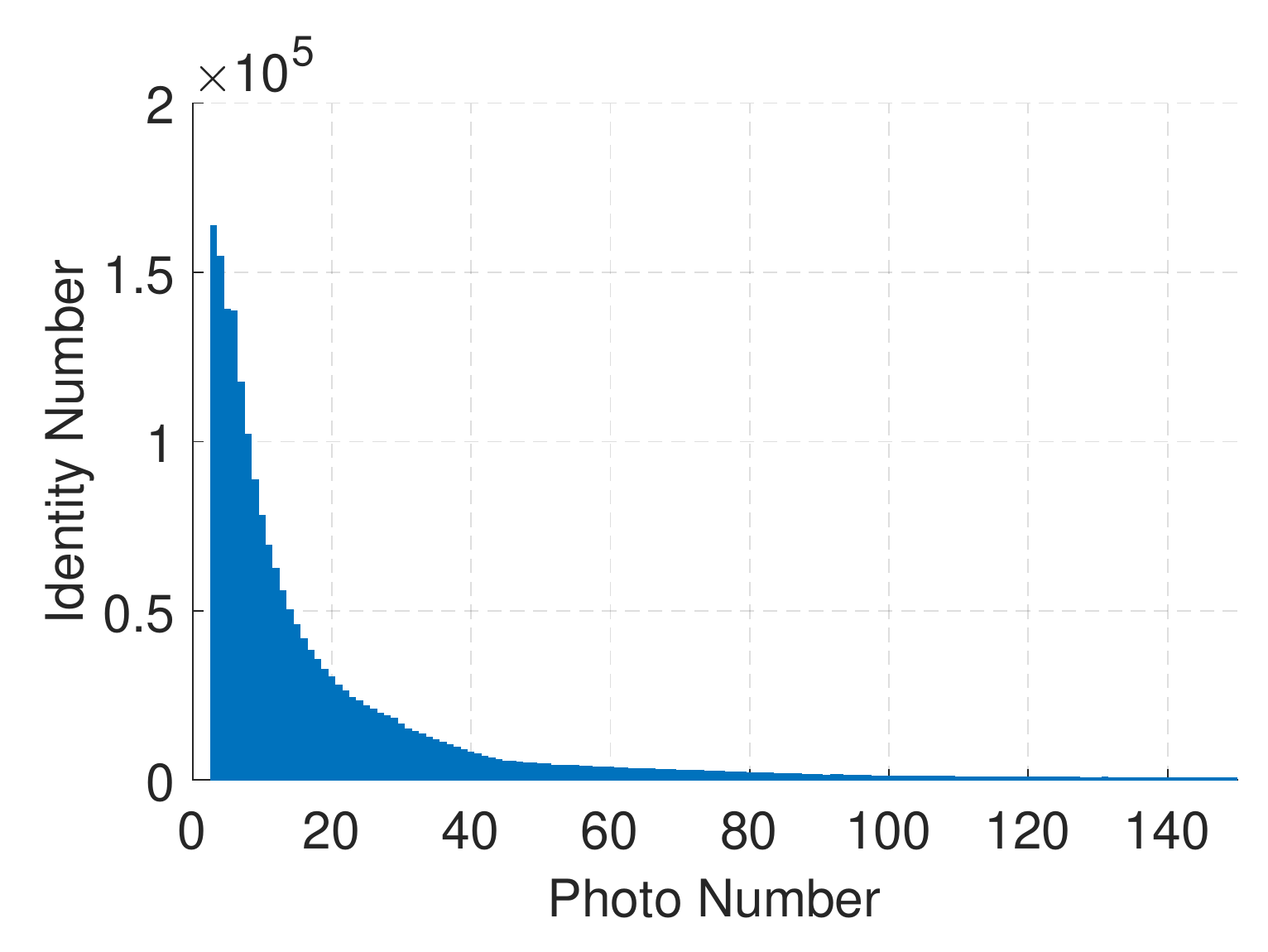}}
\vspace{-2mm}
\caption{Inter-class conflict and long-tail distribution of WebFace42M \cite{zhu2021webface260m}.}
\vspace{-4mm}
\label{fig:webfaceproblem}
\end{figure}

\begin{figure}
\centering
\subfloat[Memory Consumption]{
\label{fig:memory}
\includegraphics[width=0.22\textwidth]{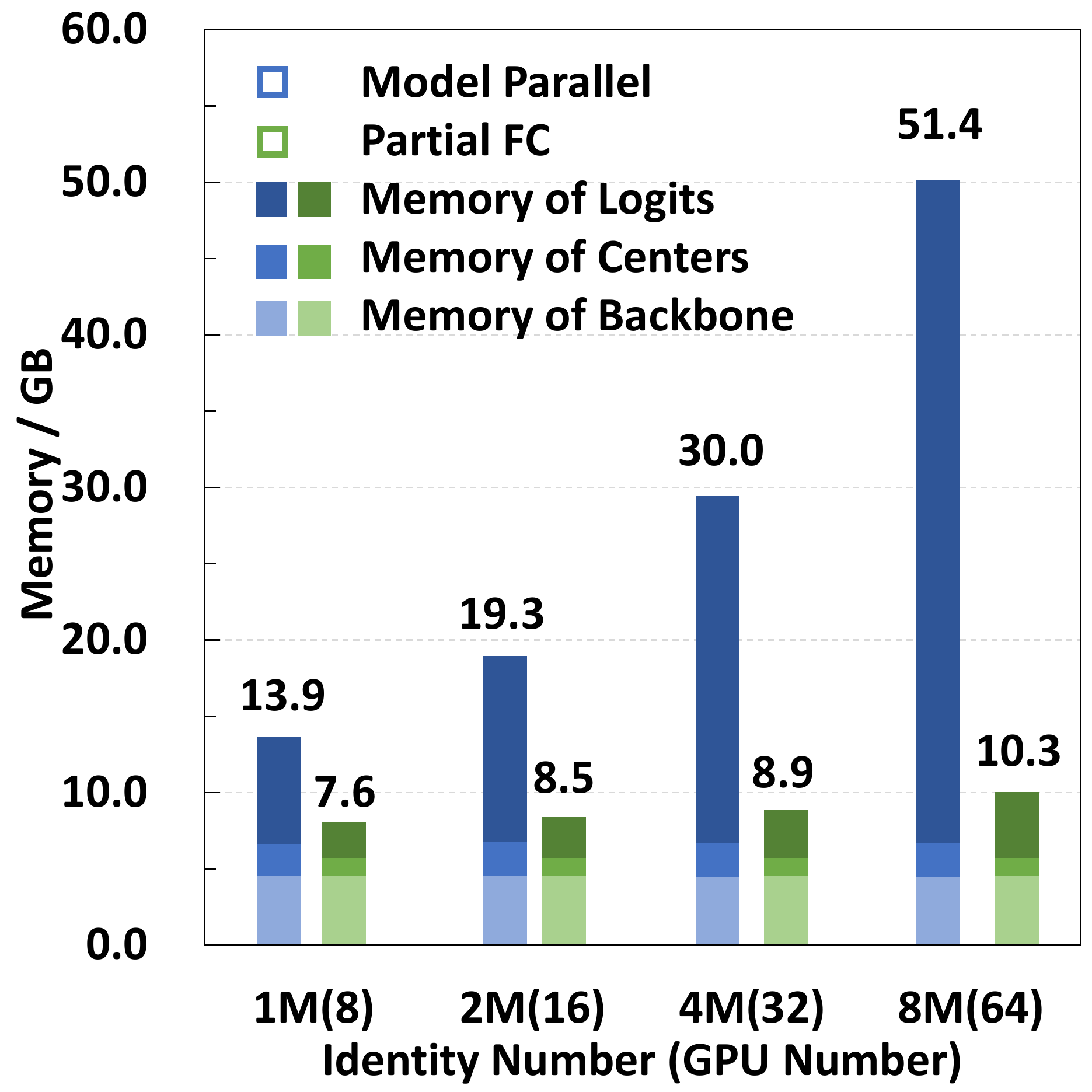}}
\subfloat[Training Speed]{
\label{fig:speed}
\includegraphics[width=0.22\textwidth]{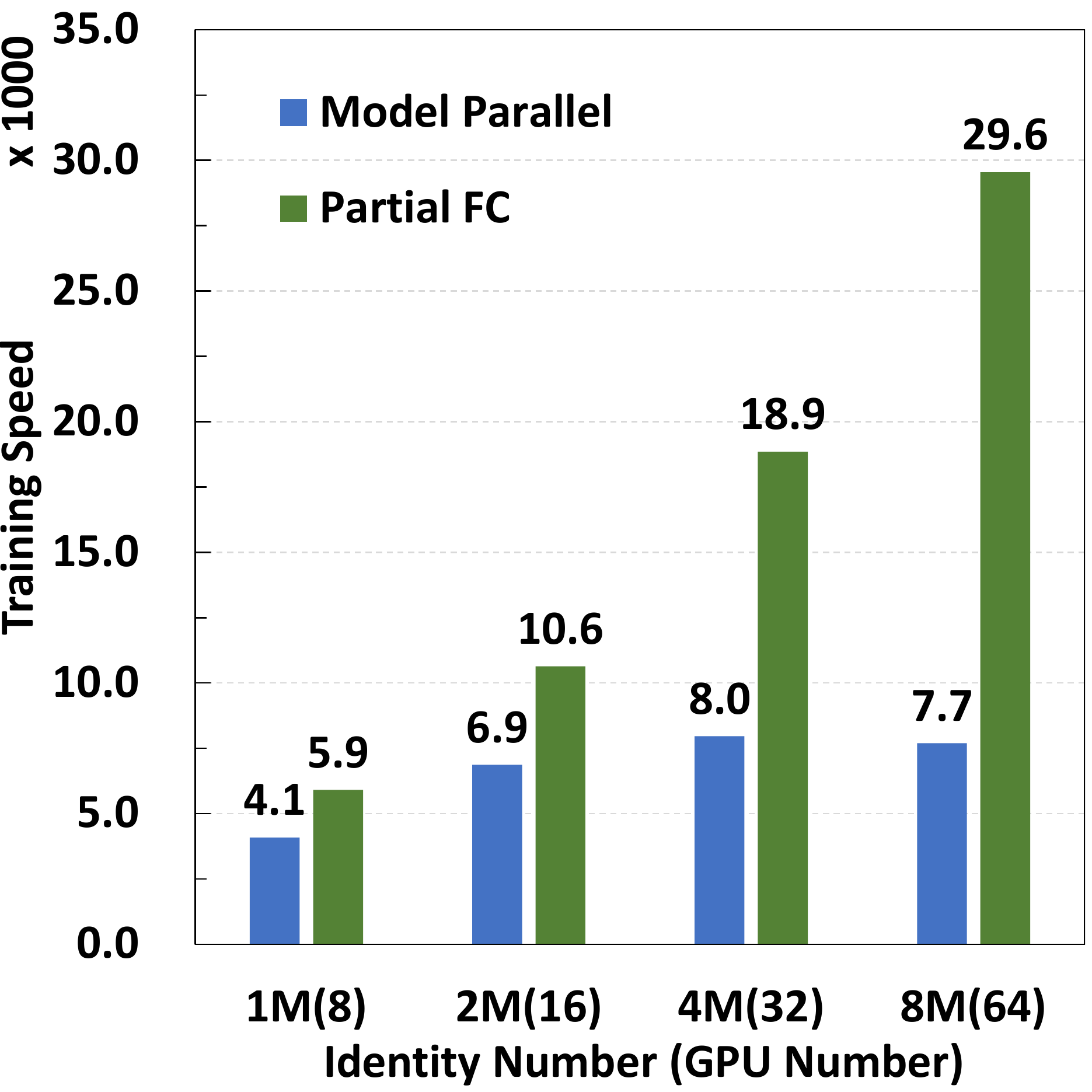}}
\vspace{-2mm}
\caption{Memory Consumption and training speed comparisons between model parallel and PFC.}
\vspace{-4mm}
\label{fig:memoryandspeed}
\end{figure}

Even though the softmax loss and its margin-based or mining-based variants achieved state-of-the-art performance on deep face recognition, the fully connected layer in the softmax loss has the following three drawbacks when applied to the large-scale web data \cite{zhu2021webface260m}.

\begin{figure*}
\centering
\includegraphics[width=0.9 \textwidth]{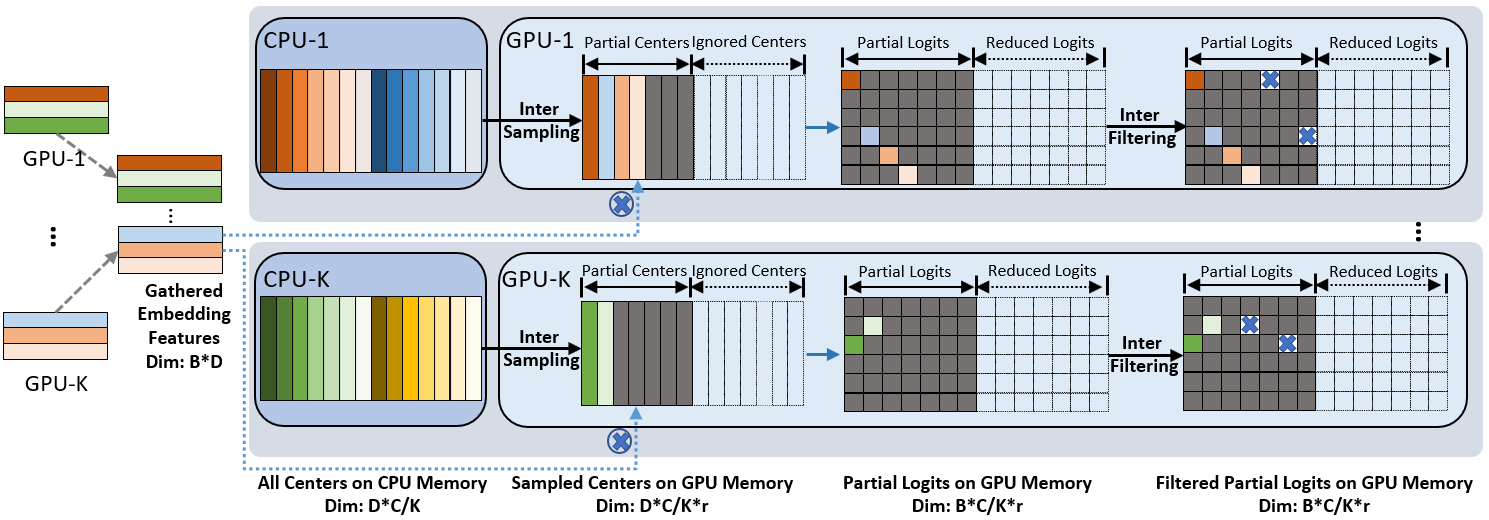}
\vspace{-2mm}
\caption{Distributed implementation of the proposed PFC. Face features are first gathered from each GPU. Meanwhile, partial centers are copied from each CPU to GPU. Positive class centers are picked through labels while partial negative class centers (in grey) are randomly selected to fill the buffer.
After the inner product between gathered features and partial centers on each GPU, we obtain the partial logits. PFC is memory-saving and efficient because it reduces the GPU memory consumption and computation cost on the FC layer. PFC is also robust under inter-class conflict due to decreased inter-class interaction during training. On the dataset with extremely heavy inter-class conflict, abnormal inter-class high similarities can be filtered by a fixed threshold (\ie 0.4) in PFC to further enhance the robustness. }
\vspace{-4mm}
\label{partial_fc}
\end{figure*}

\noindent{\bf The first limitation is the gradient confusion under inter-class conflict.}
As shown in Fig.~\ref{fig:webfaceinterclassconflict}, there are many class pairs from WebFace42M \cite{zhu2021webface260m} showing high cosine similarities (\eg $>0.4$), indicating inter-class conflict still exists in this automatically cleaned web data. Here, inter-class conflict refers to images of one person being wrongly distributed to different classes. 
If a large number of conflicted classes exist, the network optimization will suffer from gradient confusion on both features and centers, as the negative class center $W_j$ in Eq.~\ref{x-gradient} and the negative features $x^{-}_{i}$ in Eq.~\ref{PL-update} could be from the positive class.

\noindent{\bf The second limitation is that centers of tail classes undergo too many passive updates.} As shown in Fig.~\ref{fig:webfacelongtail}, the identities of WebFace42M \cite{zhu2021webface260m} are long-tail distributed, 44.57\% of identities containing less than 10 images. Under the training scenario of million-level identities and thousand-level batch size, $\mathbb{B}^+$ in Eq. \ref{PL-update} is empty in most iterations for a particular class $j$, especially for tail classes. When there is an inter-class penalty from training samples of other classes, $W_j$ is pushed away from the features of these negative samples, gradually drifting from the direction of its representing class \cite{du2020semi}. Therefore, there may exist a discrepancy between the class-wise feature center predicted by the embedding network and the corresponding center updated by SGD. 

\noindent{\bf The third limitation is that the storage and calculation of the FC layer can easily exceed current GPU capabilities.} In ArcFace \cite{deng2019arcface}, the center matrix $W\in\mathbb{R}^{D\times C}$ is equally partitioned onto $K$ GPUs.
During the forward step, each GPU first gathers all embedding features (\ie $X\in\mathbb{R}^{D\times B}$) from all GPUs.
Then, sample-to-class similarities and their exponential mappings are calculated independently on each GPU. 
To calculate the denominator of Eq. ~\ref{eq:softmax} to normalize all similarity values, the local sum on each GPU is calculated, and then the global sum is computed through cross-GPU communication.
Finally, the normalized probabilities are used in Eq.~\ref{x-gradient} and Eq.~\ref{PL-update} to calculate the feature's gradient and center's gradient.
Even though the model parallelization can completely solve the storage problems of $W$ through adding more GPUs at negligible communication cost, the storage of predicted logits can not be easily solved by increasing the GPU number.

As illustrated in Fig.~\ref{fig:memory}, the identity number $C$ goes up from 1M to 8M, and we accordingly increase the GPU number $K$ from $8$ to $64$ to keep $C/K$ consistent. However, the memory consumption of logits ($C/K\times B$) still significantly increase as the batch size rises synchronously with $K$, even surpassing the memory usage of the backbone. Besides the memory consumption of the FC layer, the computation cost in the forward and backward steps is also tremendous. As shown in Fig.~\ref{fig:speed}, the throughput can not be improved as the increased GPUs are used for the calculation of the enlarged FC layer. Therefore, simply stacking more GPUs can not efficiently solve large-scale face recognition training.

\subsection{Partial FC}
To alleviate the drawbacks of the FC layer, we propose PFC, a sparsely updating variant of the fully connected layer for training large-scale face recognition models. 
As illustrated in Fig.~\ref{partial_fc}, we maintain all of the class centers during training but randomly sample a small part of negative class centers to calculate the margin-based softmax loss instead of using all of the negative class centers in every iteration. More specifically, face feature embeddings and labels are first gathered from each GPU, and then the combined features and labels are distributed to all GPUs. In order to equalize the memory usage and computation cost of each GPU, we set a memory buffer for each GPU. The size of the memory buffer is decided by the total number of classes and the sampling rate of the negative class centers. On each GPU, positive class centers are first picked through labels and put in the buffer, then a small part of negative class centers are randomly selected to fill the rest of the buffer to ensure load balance. After the inner product between gathered embeddings and the partial center matrix on each GPU, we simultaneously obtain all partial similarity matrices to calculate the margin-based softmax loss. 

In PFC, the network will be updated towards a direction making the feature $x_i$ close to the positive class center $W^+$ and away from part of negative class centers $W^-_j$.
\begin{gather}
\frac{\partial \mathcal{L}}{\partial {x}_{i}} = -((1-p^+)W^+ - \sum_{j\in \mathbb{S},j\ne y_i}   p^-_j W^-_j),  
\label{x-gradientpfc} 
\vspace{-4mm}
\end{gather}
where $\mathbb{S}$ is a subset of all negative classes and one positive class, $\left |\mathbb{S}  \right |= C*r$, and $r$ is the sampling ratio. 
By comparing Eq.~\ref{x-gradient} and Eq.~\ref{x-gradientpfc}, we can easily find that PFC directly decreases the possibility of inter-class conflict by $r$. In addition, only positive centers and part of negative centers will be updated by Eq. \ref{PL-update} in each iteration.
Therefore, the frequency of gradient update on $W_j$ also decreases from $1.0$ to $r$, thus avoiding excessive passive update on tail class centers. In Fig.~\ref{fig:memory} and Fig.~\ref{fig:speed}, PFC saves a large amount of GPU memory used by softmax logits, thus the model training can benefit from stacking more GPUs to increase the throughput on large-scale data. 

\begin{figure}
\centering
\subfloat[Positive Cosine]{
\label{fig:positivecosine}
\includegraphics[width=0.225\textwidth]{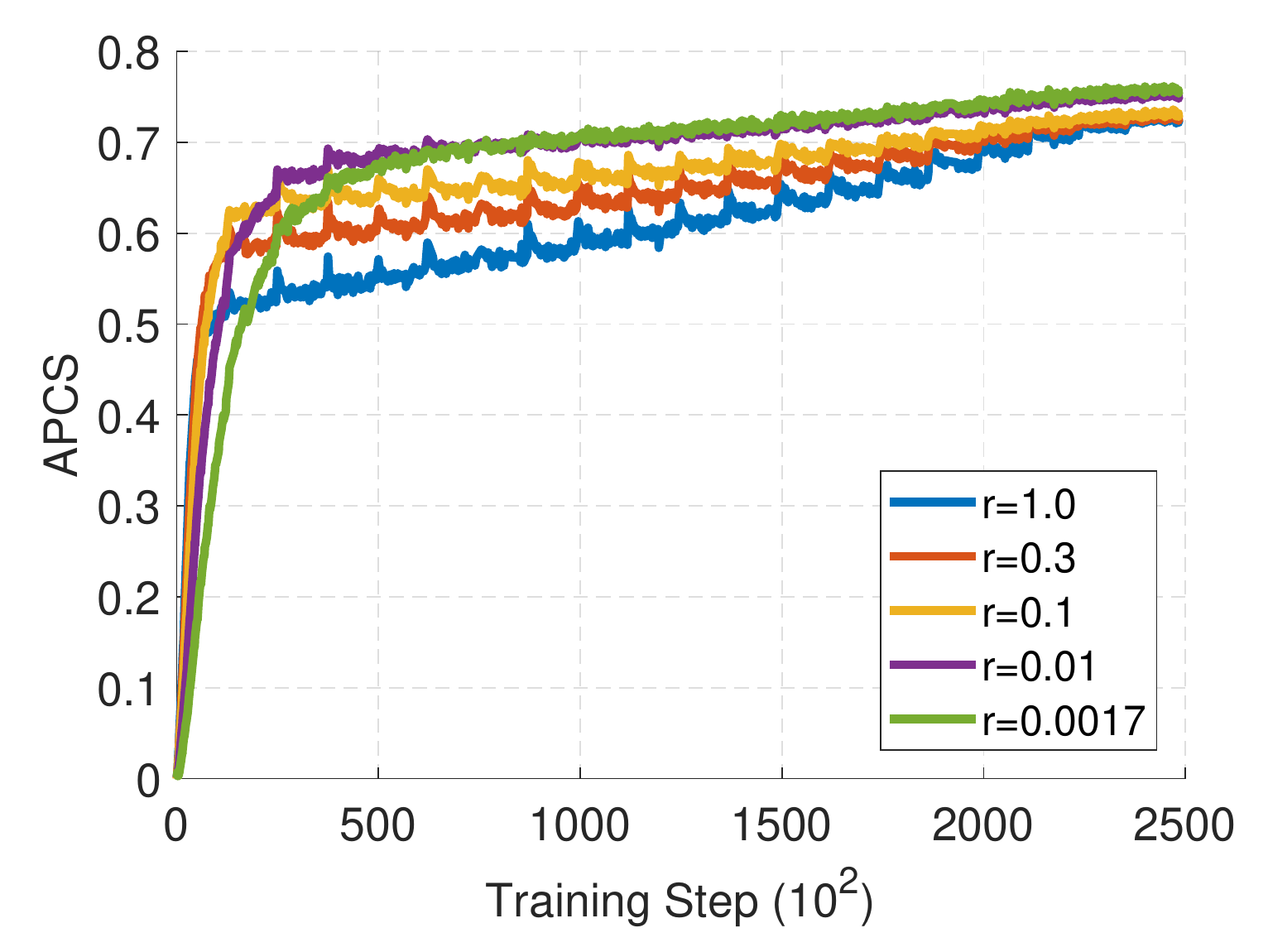}}
\subfloat[Max Cosine between Centers]{
\label{fig:maxcentercosine}
\includegraphics[width=0.225\textwidth]{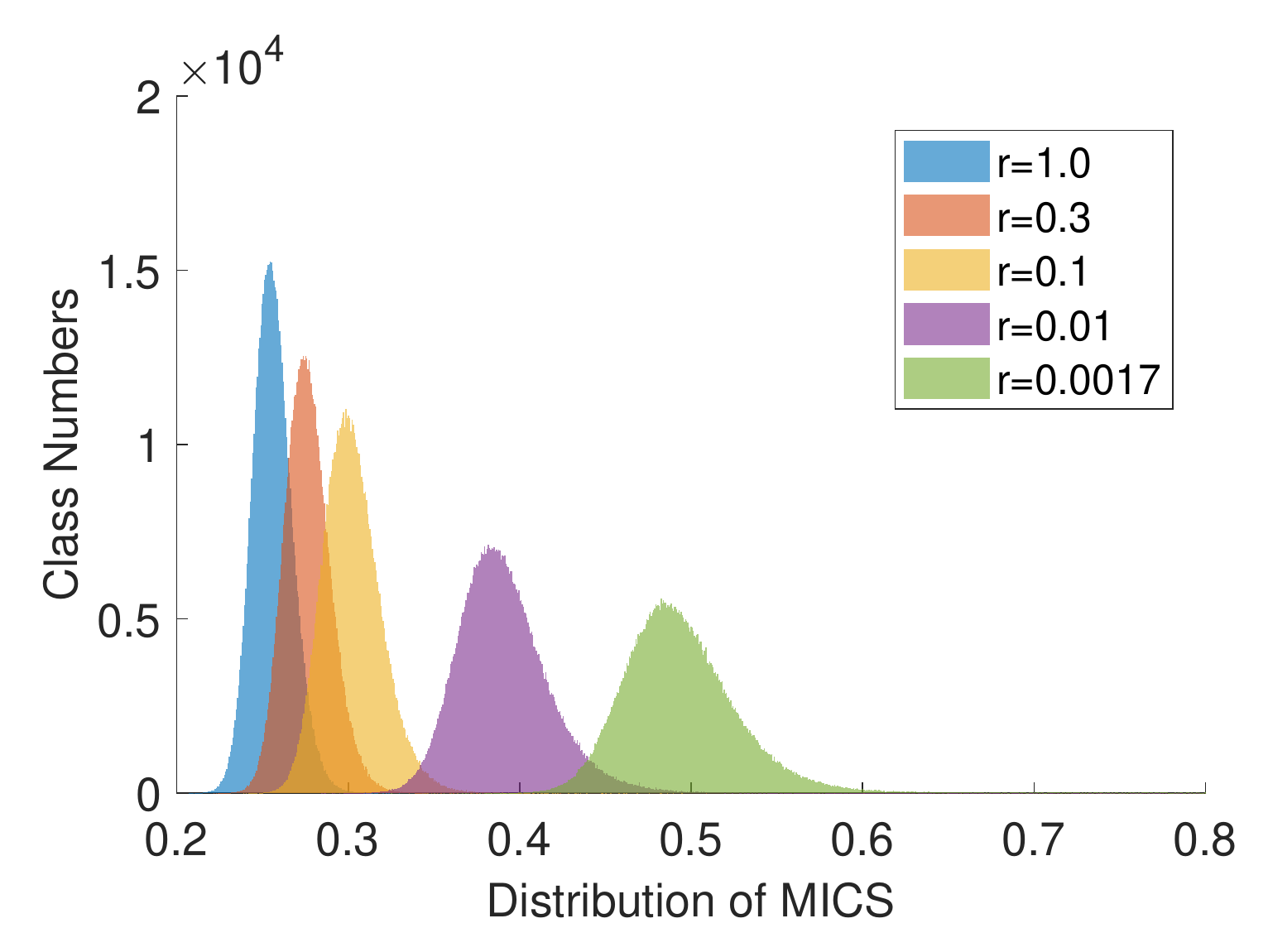}}
\vspace{-2mm}
\caption{Intra-class compactness and inter-class discrepancy comparisons under different sampling ratios on the WebFace12M dataset.}
\vspace{-4mm}
\label{fig:samplingproblem}
\end{figure}

\begin{figure}
\centering
\subfloat[IJB-C (TAR@FAR=1e-5)]{
\label{fig:ijbc}
\includegraphics[width=0.225\textwidth]{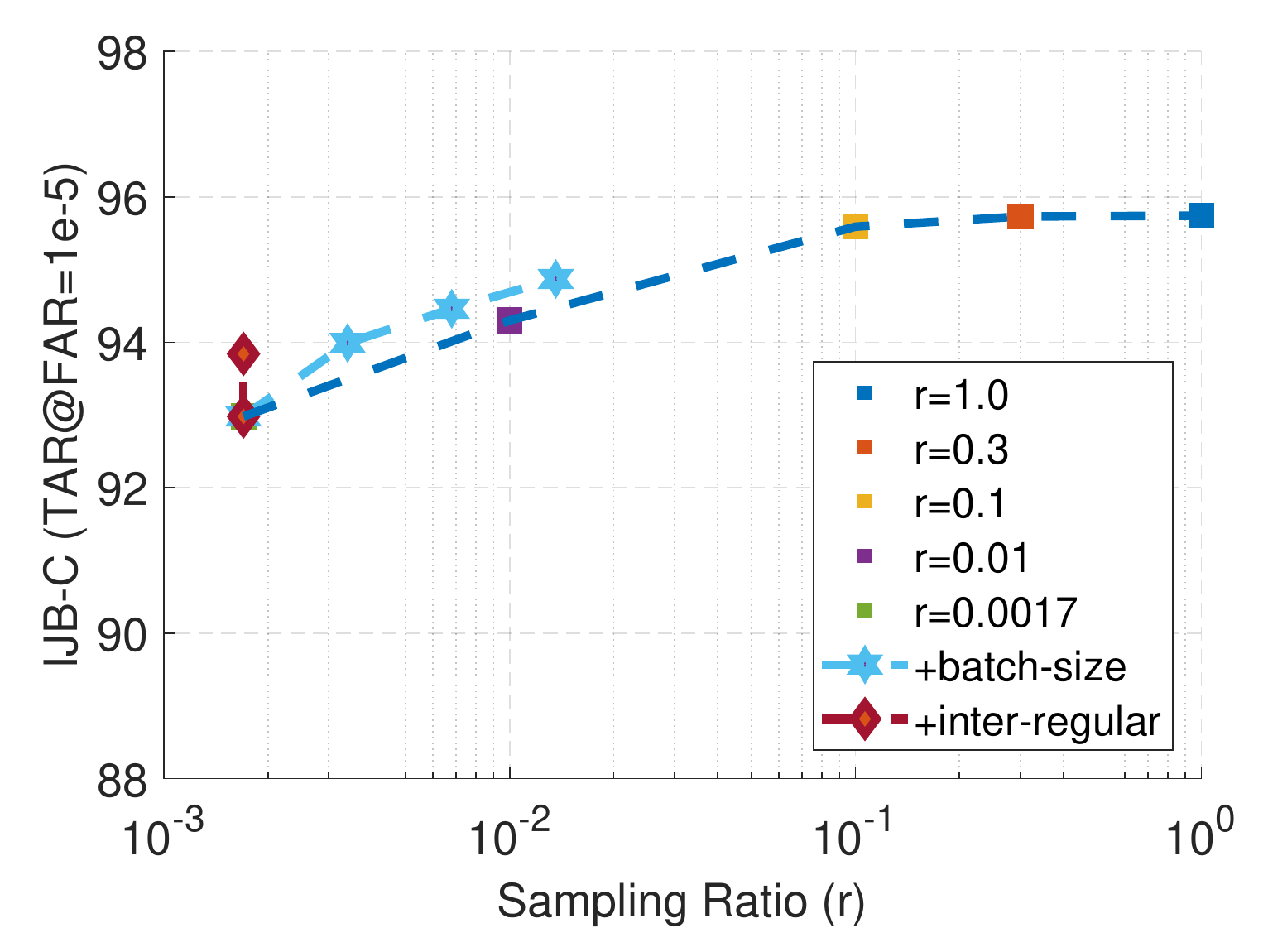}}
\subfloat[MFR-All (TAR@FAR=1e-6)]{
\label{fig:mfrall}
\includegraphics[width=0.225\textwidth]{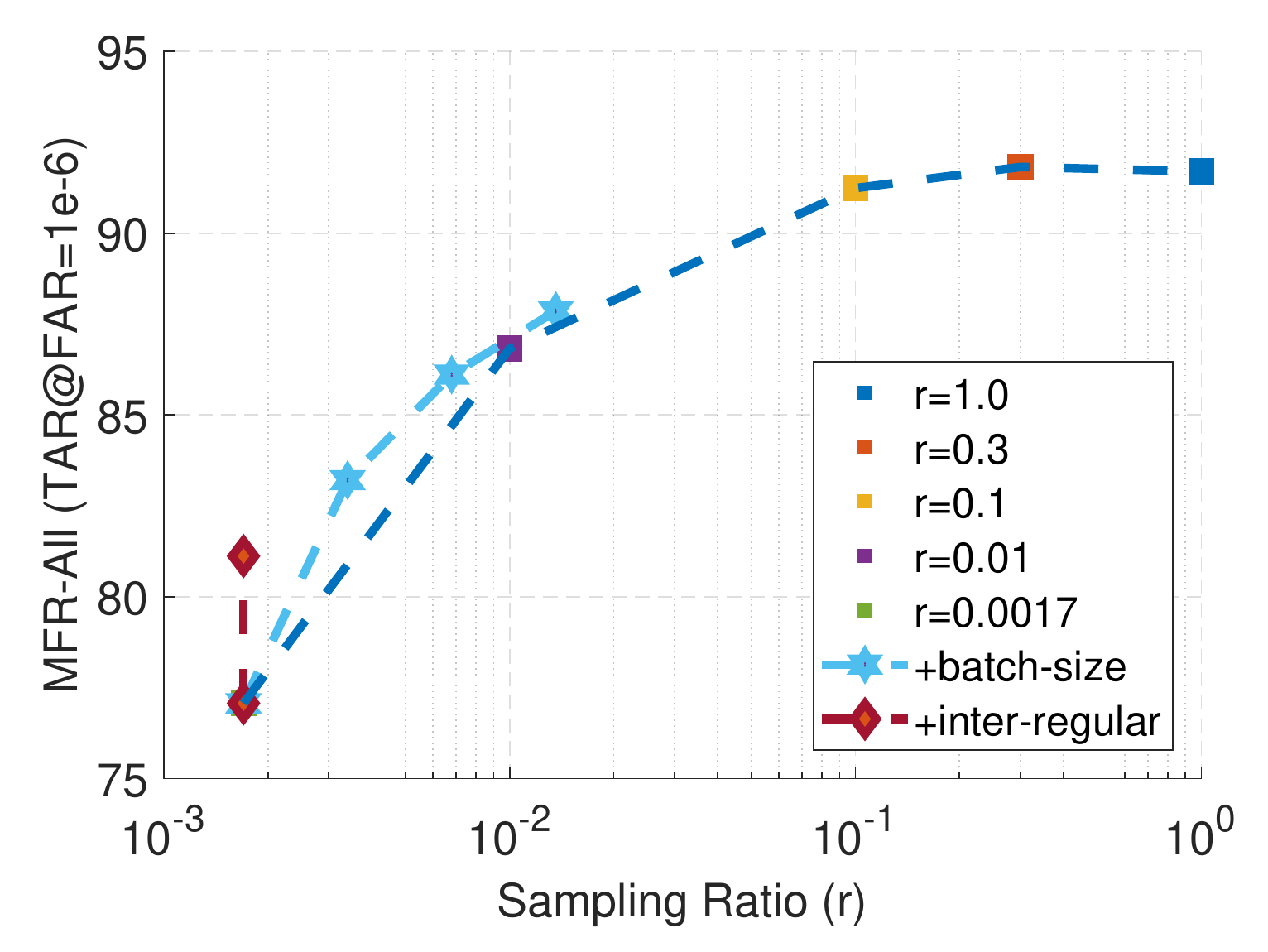}}
\vspace{-2mm}
\caption{Verification accuracy on IJB-C and MFR-All under different sampling ratios.}
\vspace{-4mm}
\label{fig:verificationacc}
\end{figure}

\begin{figure}[h]
\centering
\subfloat[Max Negative Cosine]{
\label{fig:negcosine}
\includegraphics[width=0.225\textwidth]{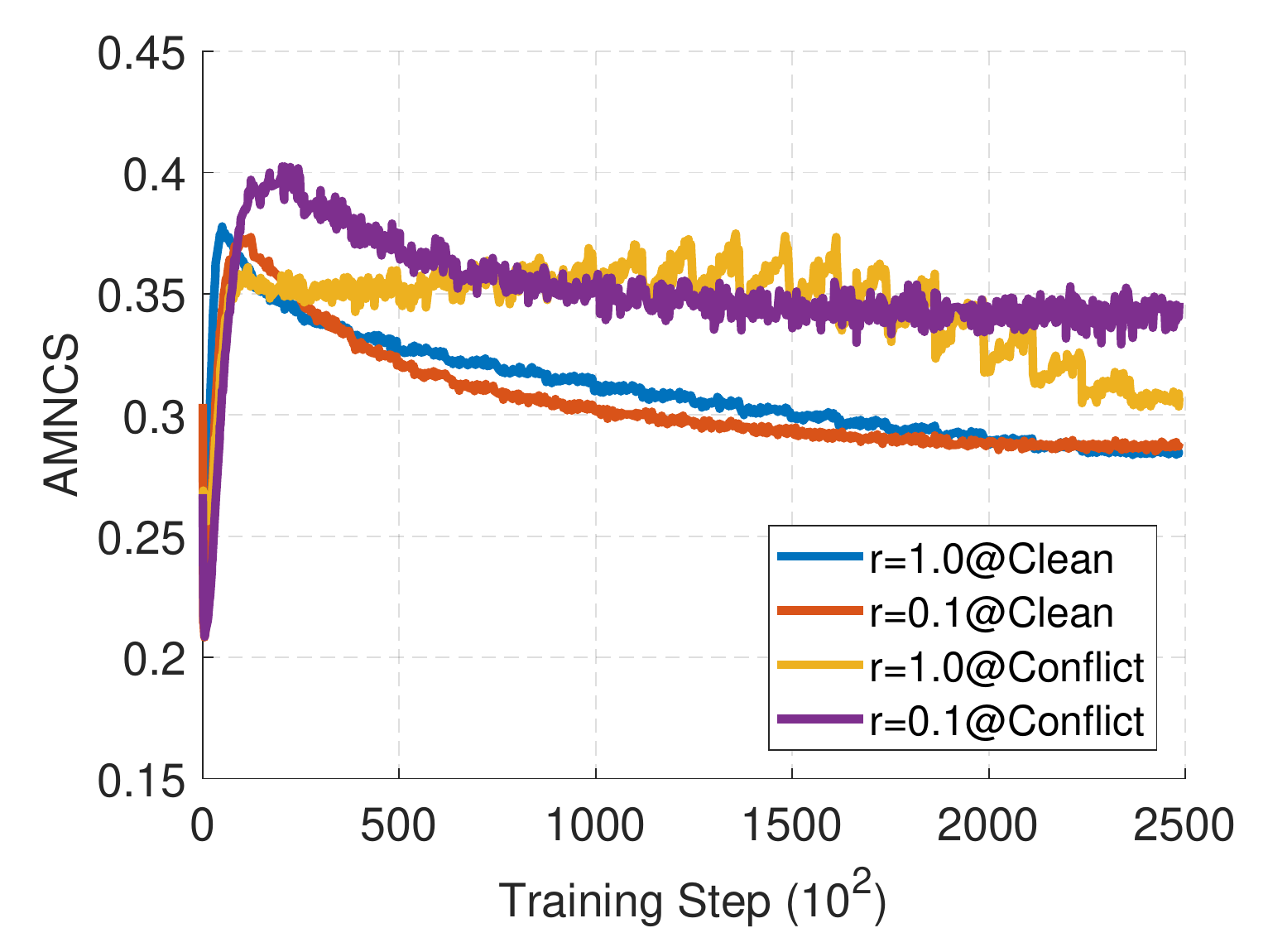}}
\subfloat[Max Cosine between Centers]{
\label{fig:maxcentercosine1}
\includegraphics[width=0.225\textwidth]{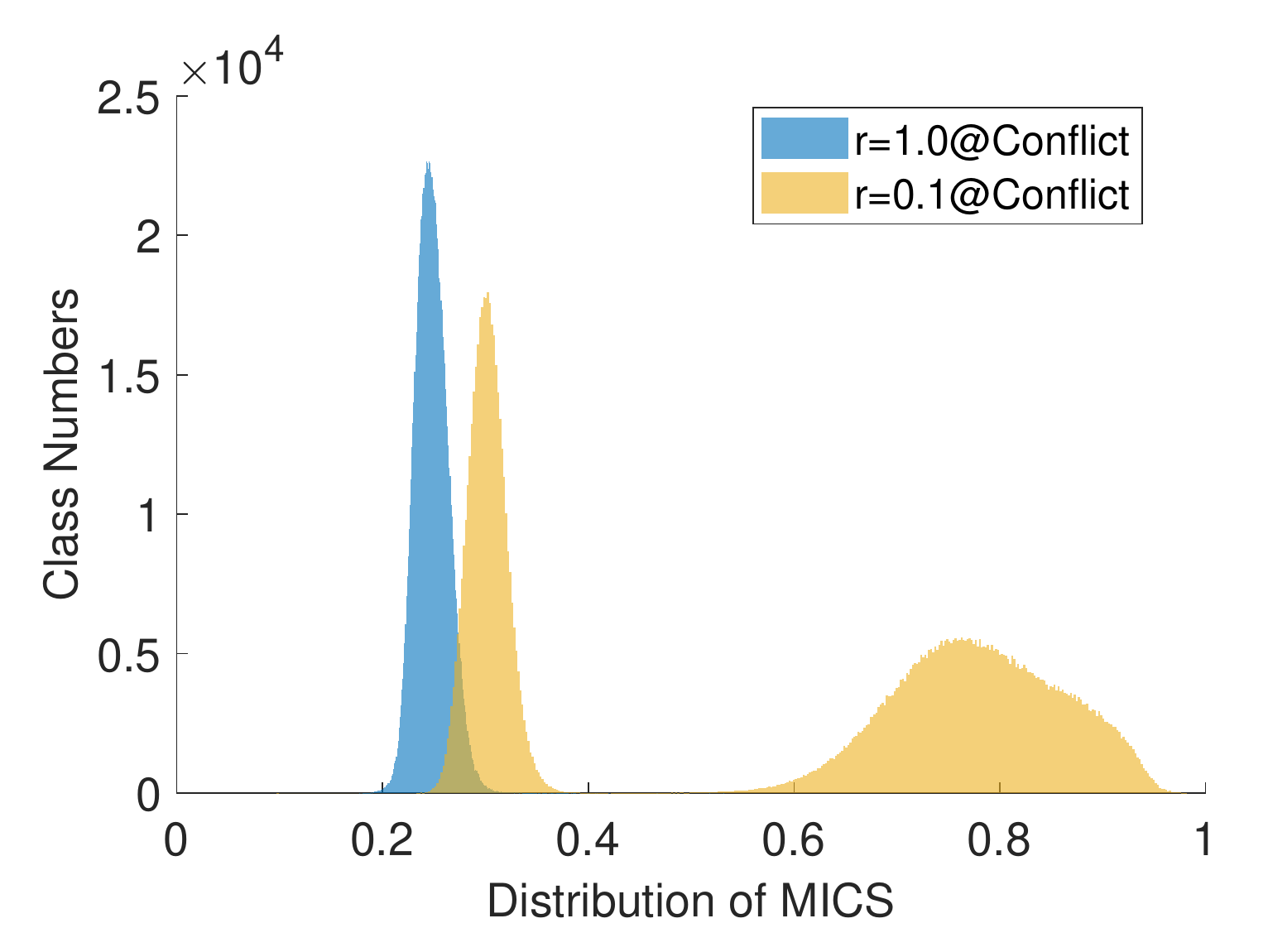}}
\vspace{-2mm}
\caption{Inter-class statistics comparisons under different sampling ratios on the WebFace12M and WebFace12M-Conflict datasets.}
\vspace{-4mm}
\label{fig:interclassnoise}
\end{figure}

\begin{figure}[h]
\centering
\subfloat[Max Negative Cosine of FC]{
\label{fig:negcosinefcsplit}
\includegraphics[width=0.225\textwidth]{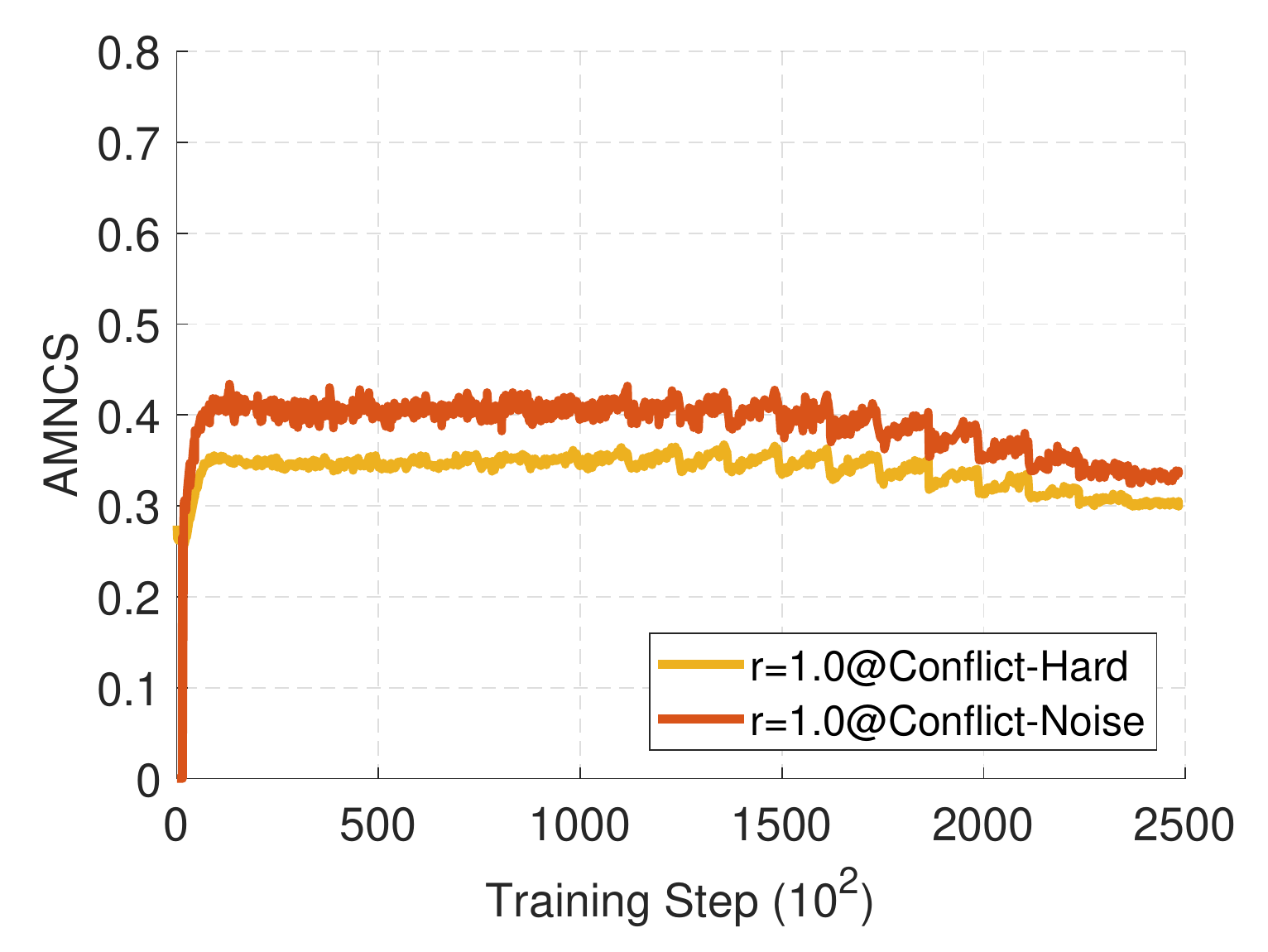}}
\subfloat[Max Negative Cosine of PFC]{
\label{fig:negcosinepfcsplit}
\includegraphics[width=0.225\textwidth]{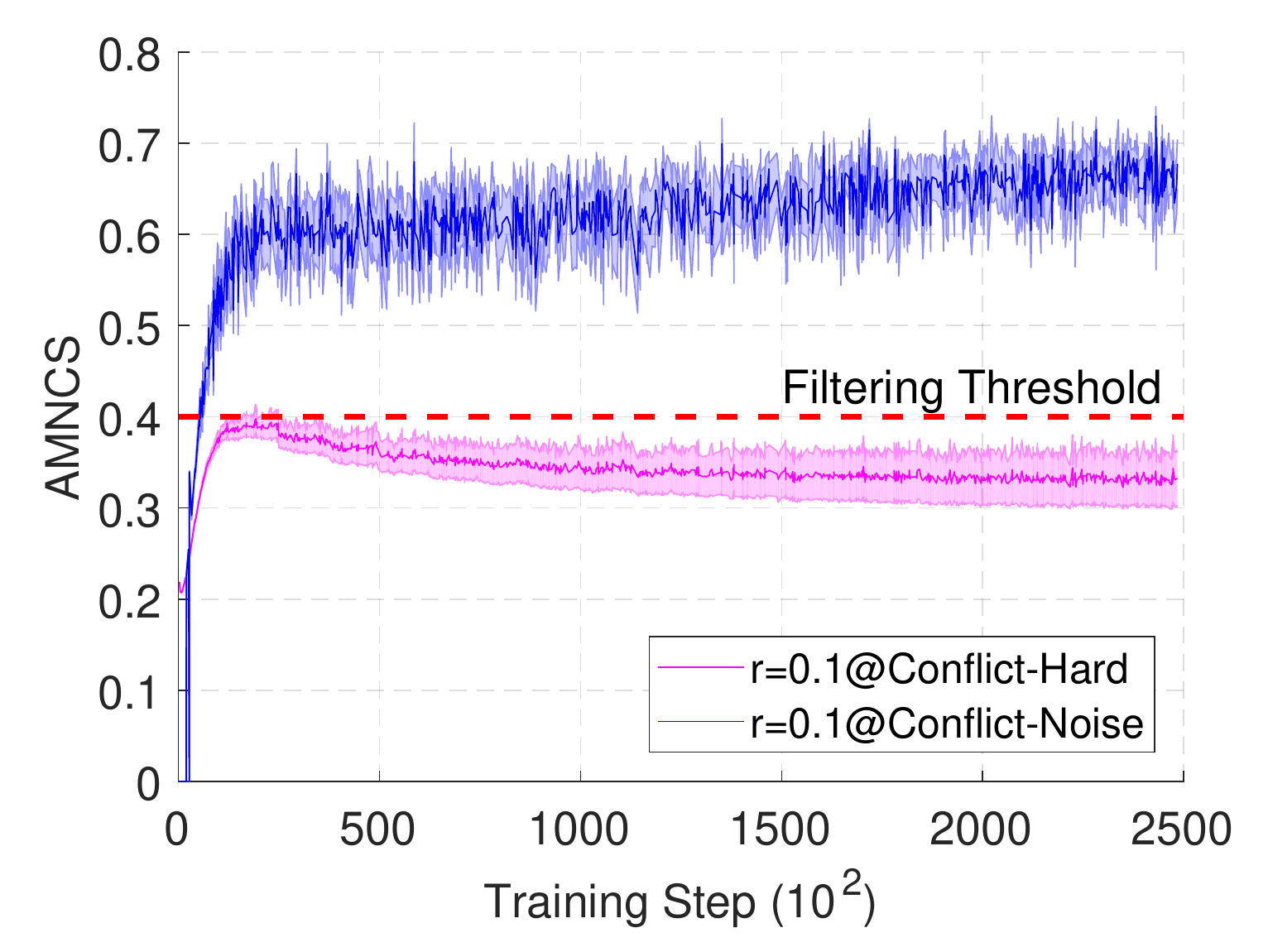}}
\vspace{-2mm}
\caption{Hard negative and conflicted negative classes analysis under different sampling ratios on the WebFace12M-Conflict datasets.}
\vspace{-4mm}
\label{fig:interclassnoisesplit}
\end{figure}

\subsection{Rethinking Inter-class Interaction}
In Eq. \ref{x-gradientpfc}, inter-class interaction between features and centers is significantly decreased by the sampling ratio.
To figure out the impact of the inter-class sampling,
we define three metrics to evaluate the real-time intra-class compactness, real-time inter-class discrepancy, and final inter-class distribution.

More specifically, we define the Average Positive Cosine Similarity (APCS) between $x_i$ and positive class center $W_{y_i}$ as $\text{APCS} = 1/B\sum_{i=1}^{B}W^T_{y_i} x_i/(\left \| W_{y_i}\right \|\left \| x_i \right \|)$ 
where $B$ is the batch-size and $\text{APCS}$ is an real-time indicator of intra-class optimization status on the training data. 
We also define the Average Maximum Negative Cosine Similarity (AMNCS) between $x_i$ and closest negative class center $W_{j}$ as $\text{AMNCS} = 1/B\sum_{i=1}^{B}\underset{j\neq i }{\text{max}}W_j^Tx_i/(\left \| W_j\right \|\left \| x_i \right \|),$
which is an real-time indicator of inter-class optimization status on the training data. To evaluate the final inter-class discrepancy, we define the Maximum Inter-class Cosine Similarity (MICS) as $\text{MICS}_i = \underset{j\neq i }{\text{max}} W_i^TW_j/(\left \| W_i\right \|\left \| W_j \right \|)$.

In Fig.~\ref{fig:samplingproblem}, we compare the intra-class and inter-class status under different sampling ratios. We train a series of ResNet50 models on the WebFace12M dataset \cite{zhu2021webface260m} by using margin-based softmax loss \cite{tencent2018CosineFace,deng2019arcface}. The minimum sampling ratio is the batch size divided by the identity number, that is $1024/600K\approx 0.0017$, indicating that only within-batch negative centers are used to construct the margin-based softmax loss. 
As illustrated in Fig.~\ref{fig:maxcentercosine}, the inter-class similarities obviously increase when the sampling ratio decreases from $1.0$ to $0.0017$. As the updating frequency of $W_j$ in Eq.~\ref{PL-update} is decreased, the network training weakens the inter-class optimization and focuses more on the intra-class optimization. Therefore, PFC achieves higher intra-class similarities during training as shown in Fig.~\ref{fig:positivecosine}. 

Even though the inter-class discrepancy deteriorates on the training data when the sampling ratio is dropping, the verification accuracy on IJB-C \cite{maze2018iarpa} and MFR-All \cite{deng2021mfrinsightface} can be still maintained when the sampling ratio is larger than $0.1$ as shown in Fig. \ref{fig:verificationacc}. When the sampling ratio drops to $0.0017$, the verification accuracy obviously decreases, indicating that inter-class interaction during training is not sufficient. To improve the inter-class discrepancy, we train three extra models by enlarging the batch size to $2K$, $4K$, $8K$ to embody more within-batch negative classes where the sampling ratio is also increased accordingly. As given in Fig. \ref{fig:verificationacc}, the performance significantly increases when the batch size is enlarged. Please note that, when PFC does not add extra negative classes outside batches, the training time consumed on the FC layer is negligible compared to the time cost on the backbone. Besides enlarging the batch size, we have also explored inter-class regularization \cite{zhao2019regularface}. By including MICS as the regularization loss, the performance can be also obviously improved in Fig. \ref{fig:verificationacc}. However, inter-class regularization needs non-ignorable computation cost on large-scale training data.

Besides the analysis on clean data, we also synthesize a WebFace12M-Conflict dataset by randomly splitting 200K identities into another 600K identities, thus WebFace12M-Conflict contains 1M pseudo classes with a high inter-class conflict ratio. As shown in Fig.~\ref{fig:negcosine}, FC ($r=1.0$) confronts with fluctuation during inter-class optimization but finally over-fits the conflicted dataset (Fig.~\ref{fig:maxcentercosine1}). By contrast, PFC ($r=0.1$) relaxes the inter-class optimization, thus conflicted classes exhibit much higher similarities as given in Fig.~\ref{fig:maxcentercosine1}. As WebFace12M-Conflict is synthesized, we can use ground-truth labels to separately calculate AMNCS for hard negative classes and conflicted negative classes (\ie inter-class noises). As shown in Fig.~\ref{fig:negcosinepfcsplit}, PFC ($r=0.1$) can punish hard negative classes as normal as on the clean dataset (Fig.~\ref{fig:negcosine}) while the conflicted negative classes can still achieve increasing similarities during training. In Fig.~\ref{fig:negcosinefcsplit}, FC ($r=1.0$) struggles to decrease the similarities between features and 
conflicted class centers, leading to over-fitting (Fig.~\ref{fig:maxcentercosine1}).
As there is a clear margin for PFC to distinguish hard negative classes and conflicted negative classes in Fig.~\ref{fig:negcosinepfcsplit}, we can further set an online inter-class filtering threshold (\ie 0.4) in PFC to suppress conflict inter-classes. In this paper, we denote PFC with abnormal inter-class filtering as PFC*.

\section{Experiments and Results}

\subsection{Implementation Details}

\noindent{\bf Datasets.}
In this paper, we employ the publicly available dataset, WebFace \cite{zhu2021webface260m}, to train face recognition models. The cleaned WebFace42M contains 2M identities, while the subsets WebFace12M and WebFace4M include 600K and 200K identities, respectively. We synthesize WebFace12M-Conflict (Tab.~\ref{ablationinterclassconflict}), WebFace12M-Flip (Tab.~\ref{ablationflipnoise}), WebFace10M-Longtail (Tab.~\ref{tab:longtail}) from WebFace to simulate the scenarios of inter-class conflict, label flip, and long-tail distribution. 

For testing, we extensively evaluate the proposed PFC on popular benchmarks, including LFW \cite{huang2007labeled}, CFP-FP \cite{sengupta2016frontal}, AgeDB \cite{Moschoglou2017AgeDB}, IJB-B \cite{whitelam2017iarpa} and IJB-C \cite{maze2018iarpa}. As the performance on these celebrity benchmarks tend to be saturated, we conduct the ablation study on MFR \cite{deng2021mfrinsightface}, which contains 1.6M images of 242K identities (non-celebrity) covering four demographic groups: African, Caucasian, South-Asian and East-Asian. On MFR, True Accept Rates (TARs) @ False Positive Rate (FAR) = 1e-6 across different races are reported from the online test server \footnote{\url{http://iccv21-mfr.com/}} after submitting the models. Besides, we also report the TARs@FAR = 1e-4 on the masked face recognition track of MFR.

\noindent{\bf Experimental settings.}
All experiments in this paper are implemented using Pytorch, and mixed-precision \cite{Micikevicius2018MixedPT} is employed to save GPU memory and accelerate training. We follow \cite{deng2019arcface,tencent2018CosineFace} to set the hyper-parameters of margin-based softmax loss and adopt flip data augmentation. We use customized ResNet \cite{he2016deep,deng2019arcface} and ViT \cite{dosovitskiy2020image} as the backbone. On different datasets, CNN models are trained for $20$ epochs while ViT models are trained for $40$ epochs.
For the training of CNN models, the default batch size per GPU is set as $128$ unless otherwise indicated. We employ the SGD optimizer with polynomial decay (power=2) and the learning rate is set as $0.1$ for the single node training (8 Tesla V100 32GB GPUs). To accelerate the training on WebFace42M, we employ four nodes with $8\times4$ GPUs and linearly warm up the learning rate from $0$ to $0.4$ within the first $2$ epochs. Afterwards, polynomial decay (power=2) is used for another 18 epochs as shown in Fig.~\ref{fig:losslr}.
For the training of ViT models, the default batch size per GPU is set as $384$. We use the AdamW \cite{loshchilov2017decoupled} optimizer with a base learning rate of $0.001$ and a weight decay of $0.1$. To achieve quick training, we employ eight nodes with $8\times8$ GPUs and linearly warm up the learning rate from $0$ to $0.001$ within the first 4 epochs. Then, polynomial decay (power=2) is used for another 36 epochs.  

\begin{table}[t]
\centering
\resizebox{\linewidth}{!}{
\begin{tabular}{ l | c | c | c | c | c | c}
\hline       
\multirow{2}{*}{Datasets} & \multicolumn{6}{c } {MFR} \cr
\cline{2-7} & \textbf{All} & Afr & Cau & S-Asian & E-Asian & \textbf{Mask} \cr
\hline   
WF4M+FC-1.0           & 86.25                               & 83.35 & 91.11 & 88.14 & 65.79 & 72.05 \cr 
WF4M+PFC-0.04         & 74.11 (- 12.14)                     & 71.43 & 81.79 & 76.25 & 52.24 & 54.21 \cr 
WF4M+PFC-0.1          & 85.76 (- 0.49)                      & 83.82 & 91.00 & 87.90 & 66.04 & 71.13 \cr 
WF4M+PFC-0.2          & 86.36 (\textcolor{Emerald}{+ 0.11}) & 84.47 & 91.39 & 88.45 & 66.61 & 71.88 \cr 
WF4M+PFC-0.3          & 86.85 (\textcolor{Emerald}{+ 0.60}) & 84.86 & 91.57 & 88.57 & 67.52 & 72.28 \cr 
WF4M+PFC-0.4          & 86.81 (\textcolor{Emerald}{+ 0.56}) & 84.75 & 91.44 & 88.41 & 67.17 & 71.99 \cr 
\hline   
WF12M+FC-1.0          & 91.70                               & 90.72 & 94.94 & 93.44 & 75.10  & 80.47 \cr 
WF12M+PFC-0.013       & 87.85 (- 3.85)                      & 87.07 & 92.32 & 90.70 & 68.28  & 72.98 \cr 
WF12M+PFC-0.1         & 91.24 (- 0.46)                      & 90.80 & 94.67 & 93.18 & 74.97  & 79.73 \cr 
WF12M+PFC-0.2         & 91.78 (\textcolor{Emerald}{+ 0.08}) & 91.09 & 95.00 & 93.53 & 75.90  & 79.92 \cr 
WF12M+PFC-0.3         & 91.82 (\textcolor{Emerald}{+ 0.12}) & 91.14 & 95.00 & 93.61 & 75.55  & 80.08 \cr 
WF12M+PFC-0.4         & 91.81 (\textcolor{Emerald}{+ 0.11}) & 90.97 & 95.03 & 93.40 & 75.55  & 80.61 \cr 
\hline
WF42M+FC-1.0          & 93.86                               & 93.33 & 96.20 & 95.24 & 79.46 & 83.90 \cr     
WF42M+PFC-0.008       & 91.27 (- 2.59)                      & 90.34 & 95.16 & 93.04 & 76.93 & 81.24 \cr 
WF42M+PFC-0.1         & 93.95 (\textcolor{Emerald}{+ 0.09}) & 93.48 & 96.37 & 95.51 & 80.03 & 83.79 \cr 
WF42M+PFC-0.2         & 94.04 (\textcolor{Emerald}{+ 0.18}) & 93.67 & 96.38 & 95.49 & 80.07 & 84.32 \cr 
WF42M+PFC-0.3         & 94.03 (\textcolor{Emerald}{+ 0.17}) & 93.68 & 96.38 & 95.52 & 79.76 & 84.46 \cr 
WF42M+PFC-0.4         & 93.95 (\textcolor{Emerald}{+ 0.09}) & 93.38 & 96.35 & 95.46 & 79.57 & 84.42 \cr 
\hline
\end{tabular}
}
\vspace{-2mm}
\caption{Performance comparisons under different sampling ratios on different training datasets. ResNet50 is used here.}
\label{ablationsampingratio}
\vspace{-4mm}
\end{table}

\begin{table}[t]
\centering
\resizebox{\linewidth}{!}{
\begin{tabular}{ l | c| c | c | c | c | c | c | c}
\hline       
\multirow{2}{*}{Net} & \multirow{2}{*}{GFlops} & IJB-C &\multicolumn{6}{c} {MFR} \cr
  \cline{3-9}   & &1e-5 &\textbf{All}  & Afr   & Cau   & S-Asian & E-Asian  & \textbf{Mask}\cr
\hline
R18             & 2.62  & 93.36 & 79.13 & 75.50 & 86.10 & 80.55 & 57.77 & 63.87 \cr 
R50             & 6.33  & 95.94 & 94.03 & 93.68 & 96.38 & 95.52 & 79.76 & 84.31 \cr  
R100            & 12.12 & 96.45 & 96.69 & 96.68 & 98.09 & 97.72 & 86.14 & 89.64 \cr  
R200            & 23.47 & 96.93 & 97.70 & 97.79 & 98.70 & 98.54 & 89.52 & 91.87 \cr  
\hline
ViT-T             &1.51  & 95.97 &  92.30 & 91.72 & 95.20 & 93.63 & 77.41 & 78.46 \cr 
ViT-S             &5.74  & 96.57 &  95.87 & 95.74 & 97.47 & 96.85 & 84.87 & 85.82 \cr  
ViT-B             &11.42 & 97.04 &  97.42 & 97.62 & 98.53 & 98.20 & 88.77 & 89.48 \cr  
ViT-L             &25.31 & 97.23 &  97.85 & 98.07 & 98.81 & 98.66 & 89.97 & 90.88 \cr  
\hline
\end{tabular}
}
\vspace{-2mm}
\caption{Performance analysis of PFC ($r$=0.3) using different network structures (\ie CNN and ViT). Here, WebFace42M is used as the training data and the gradient check-pointing \cite{chen2016training} is used to save memory.}
\label{ablationdifferentbackbone}
\vspace{-4mm}
\end{table}

\begin{table}[t]
\centering
\resizebox{\linewidth}{!}{
\begin{tabular}{ l | c | c | c | c | c }
\hline       
\multirow{2}{*}{Datasets} & \multicolumn{5}{c} {MFR} \cr
\cline{2-6} & \textbf{All}  & Afr   & Cau   & S-Asian & E-Asian \cr
\hline
WF12M+FC-1.0               & 91.70 & 90.72 & 94.94 & 93.44 & 75.10  \cr 
WF12M-Conflict+FC-1.0      & 79.93 & 79.09 & 87.56 & 84.49 & 55.83  \cr  
WF12M-Conflict+FC*-1.0     & 91.18 & 90.28 & 94.52 & 92.74 & 74.37  \cr  
\hline
WF12M-Conflict+PFC-0.1      & 91.20 (\textcolor{Emerald}{+ 11.27}) & 90.65 & 94.65 & 93.40 & 74.99  \cr  
WF12M-Conflict+PFC*-0.1     & 91.58 (\textcolor{Emerald}{+ 11.65}) & 91.01 & 94.81 & 93.42 & 75.42  \cr 
\hline
WF12M-Conflict+PFC-0.2      & 90.55 (\textcolor{Emerald}{+ 10.62}) & 90.43 & 94.33 & 93.13 & 73.53  \cr  
WF12M-Conflict+PFC*-0.2     & 91.68 (\textcolor{Emerald}{+ 11.75}) & 91.19 & 95.04 & 93.64 & 75.52  \cr  
\hline
WF12M-Conflict+PFC-0.3      & 89.59 (\textcolor{Emerald}{+ 9.66})  & 89.24 & 93.67 & 92.35 & 71.85  \cr 
WF12M-Conflict+PFC*-0.3     & 91.68 (\textcolor{Emerald}{+ 11.75}) & 91.03 & 94.85 & 93.60 & 75.51  \cr  
\hline
WF12M-Conflict+PFC-0.4      & 87.78 (\textcolor{Emerald}{+ 7.85})  & 87.51 & 92.63 & 91.04 & 68.59  \cr  
WF12M-Conflict+PFC*-0.4     & 91.54 (\textcolor{Emerald}{+ 11.61}) & 91.07 & 94.63 & 93.57 & 75.48  \cr  
\hline
\end{tabular}
}
\vspace{-2mm}
\caption{Performance analysis of PFC under synthetic inter-class conflict. The WebFace12M-Conflict dataset contains 1M classes split from the WebFace12M dataset. ResNet50 is used here. ``+PFC*'' denotes additional inter-class filtering to neglect abnormal negative class centers with cosine similarities higher than $0.4$.}
\label{ablationinterclassconflict}
\vspace{-4mm}
\end{table}

\subsection{Ablation Study}
\noindent{\bf PFC across different datasets and sampling ratios.}In Tab.~\ref{ablationsampingratio}, we train ResNet50 models on three different datasets with different sampling ratios. Compared to the performance of FC, PFC-$0.1$ not only accelerates the training but also achieves comparable results across different datasets with identities ranging from $200K$ to $2M$. When the sampling ratio is increased to $0.2$ and $0.3$, PFC exhibits consistent better performance than the baseline, indicating random sampling on large-scale datasets is beneficial for both training speed and model's robustness. When the sampling ratio is too small (\eg around $0.01$), there is an obvious performance drop on MFR-All because (1) the inter-class interaction is insufficient under a low sampling ratio during training, and (2) trillion-level negative pair comparison during testing is very challenging.

\noindent{\bf PFC across different network structures.}In Tab.~\ref{ablationdifferentbackbone}, we train PFC ($r$=0.3) on the WebFace42M dataset by using CNN or ViT as the backbone. As can be seen, PFC achieves impressive performance across different network complexities and ViT-based networks can obtain better performance than CNN-based networks under the similar computation cost. Specifically, the ViT large (ViT-L) model obtains $97.23\%$ TAR@FAR =1e-5 on IJB-C and $97.85\%$ TAR@FAR =1e-6 on MFR-All.

\begin{table}[t]
\centering
\resizebox{\linewidth}{!}{
\begin{tabular}{ l | c | c | c | c | c }
\hline       
\multirow{2}{*}{Datasets} & \multicolumn{5}{c } {MFR} \cr
\cline{2-6} & \textbf{All} & Afr & Cau & S-Asian & E-Asian \cr
\hline
WF12M+FC-1.0         & 91.70                               & 90.72 & 94.94 & 93.44 & 75.10 \cr
WF12M+PFC-0.1        & 91.24 (- 0.46)                      & 90.80 & 94.67 & 93.18 & 74.97 \cr 
WF12M+PFC*-0.1       & 91.53 (- 0.17)                      & 90.99 & 94.87 & 93.34 & 75.26 \cr 
\hline
WF12M-Flip(10\%)+FC-1.0      & 88.77 & 87.12 & 92.81 & 90.58 & 70.74 \cr      
WF12M-Flip(10\%)+PFC-0.1  & 89.60 (\textcolor{Emerald}{+ 0.83})& 89.60 & 94.02 & 92.19 & 72.23 \cr 
WF12M-Flip(10\%)+PFC*-0.1 & 90.03 (\textcolor{Emerald}{+ 1.26})& 89.80 & 94.12 & 92.25 & 73.27 \cr  
\hline
WF12M-Flip(20\%)+FC-1.0      & 85.42 & 83.98 & 90.92 & 87.95 & 65.54 \cr      
WF12M-Flip(20\%)+PFC-0.1  & 87.62 (\textcolor{Emerald}{+ 2.20})& 87.53 & 92.82 & 90.87 & 69.27 \cr
WF12M-Flip(20\%)+PFC*-0.1 & 88.17 (\textcolor{Emerald}{+ 2.75})& 87.96 & 93.20 & 91.15 & 70.05 \cr 
\hline
WF12M-Flip(40\%)+FC-1.0   & 43.87 & 41.61 & 52.80 & 48.03 & 28.60 \cr      
WF12M-Flip(40\%)+PFC-0.1  & 78.53 (\textcolor{Emerald}{+ 34.66})& 79.33 & 87.52 & 83.86 & 57.54 \cr 
WF12M-Flip(40\%)+PFC*-0.1 & 80.20 (\textcolor{Emerald}{+ 36.33})& 80.57 & 88.66 & 85.03 & 59.94 \cr 
\hline
\end{tabular}
}
\vspace{-2mm}
\caption{Performance analysis of PFC under different synthetic label-flip noise ratios (\eg 10\%, 20\% and 40\%). ResNet50 is used here. ``+PFC*'' denotes additional abnormal inter-class filtering as in Tab.~\ref{ablationinterclassconflict}.}
\label{ablationflipnoise}
\vspace{-4mm}
\end{table}

\begin{table}[t]
\centering
\resizebox{\linewidth}{!}{
\begin{tabular}{ l | c | c | c | c | c }
\hline       
\multirow{2}{*}{Datasets} & \multicolumn{5}{c } {MFR} \cr
\cline{2-6} & \textbf{All} & Afr & Cau & S-Asian & E-Asian \cr
\hline
WF10M-Longtail-FC-1.0            & 87.44 & 85.79 & 91.86 & 89.30 & 69.39 \cr 
WF10M-Longtail-DCQ               & 89.37 & 87.24 & 92.16 & 91.94 & 71.35 \cr
WF10M-Longtail-PFC-0.1           & 91.92 (\textcolor{Emerald}{+ 4.48}) & 90.73 & 94.80 & 92.77 & 76.18 \cr 
WF10M-Longtail-PFC-0.2           & 91.96 (\textcolor{Emerald}{+ 4.52}) & 91.14 & 95.05 & 93.54 & 76.53 \cr 
WF10M-Longtail-PFC-0.3           & 91.64 (\textcolor{Emerald}{+ 4.20}) & 90.58 & 94.75 & 93.28 & 76.02 \cr 
WF10M-Longtail-PFC-0.4           & 91.03 (\textcolor{Emerald}{+ 3.59}) & 90.11 & 94.57 & 93.07 & 75.98 \cr 
\hline
\end{tabular}
}
\vspace{-2mm}
\caption{Performance analysis of PFC models trained on the WebFace10M-Longtail dataset. ResNet50 is used here.}
\label{tab:longtail}
\vspace{-4mm}
\end{table}

\begin{table}
\centering
\resizebox{\linewidth}{!}{
\begin{tabular}{l | c |c | c|c | c| c}
\hline
Method         &  ID  &  GPU & BS &  Mem & Speed & MFR-All   \\
\hline
Model Parallel &  2M   & 8   & 128 & 18.9  & 2463   & 95.35  \\
HF-Softmax     &  2M   & 8   & 128 & 10.7  & 1034   & 93.21  \\
D-Softmax      &  2M   & 8   & 128 & 13.8  & 1840   & 91.69  \\
PFC-0.1        &  2M   & 8   & 128 & 11.8  & 3552   & 96.19  \\ \hline
Model Parallel &  10M  & 8   & 16  & 32.0  & 502    & -      \\
PFC-0.1        &  10M  & 8   & 64  & 14.1  & 2497   & -      \\ \hline
Model Parallel &  2M   & 64  & 240 & 30.5  & 15357  & 95.46  \\
PFC-0.1        &  2M   & 64  & 240 & 17.2  & 23396  & 96.08  \\ \hline
Model Parallel &  10M  & 64  & 72  & 27.2  & 4840   & -      \\
PFC-0.1        &  10M  & 64  & 72  & 9.4   & 17819  & -      \\
\hline
\end{tabular}
}
\vspace{-2mm}
\caption{Large-scale training comparison on WebFace42M and synthetic 10M identities. ResNet100 and V100 GPUs are used here. ``BS'' abbreviates the batch size. The memory is GPU storage in GB and the speed is throughput in samples/second.}
\vspace{-4mm}
\label{tab:momandefficiency}
\end{table}

\begin{figure}
\centering
\subfloat[Loss\&LR]{
\label{fig:losslr}
\includegraphics[height=0.4\linewidth]{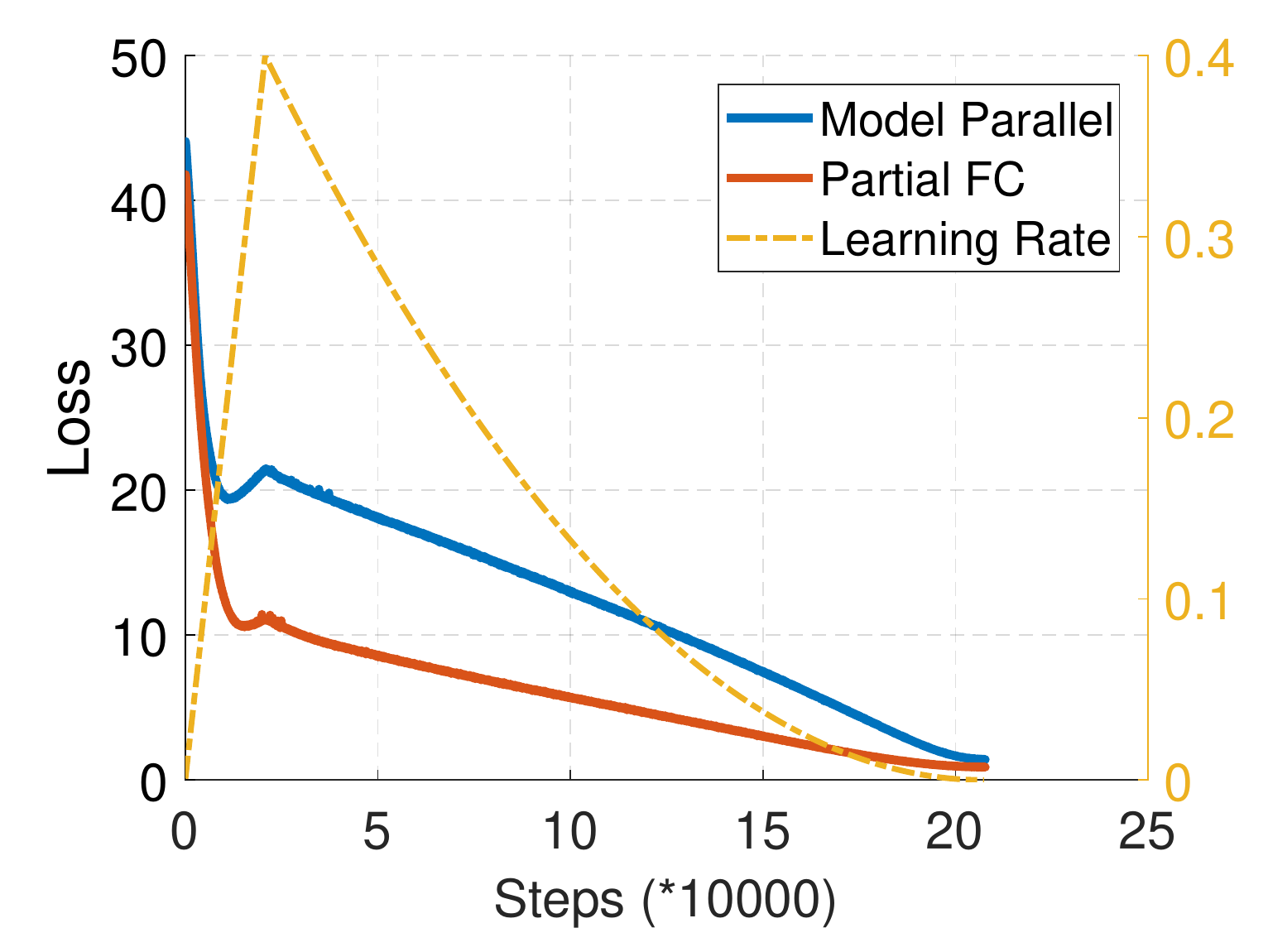}}
\subfloat[Ver. on MFR-All]{
\label{fig:webface2mspeed}
\includegraphics[height=0.39\linewidth]{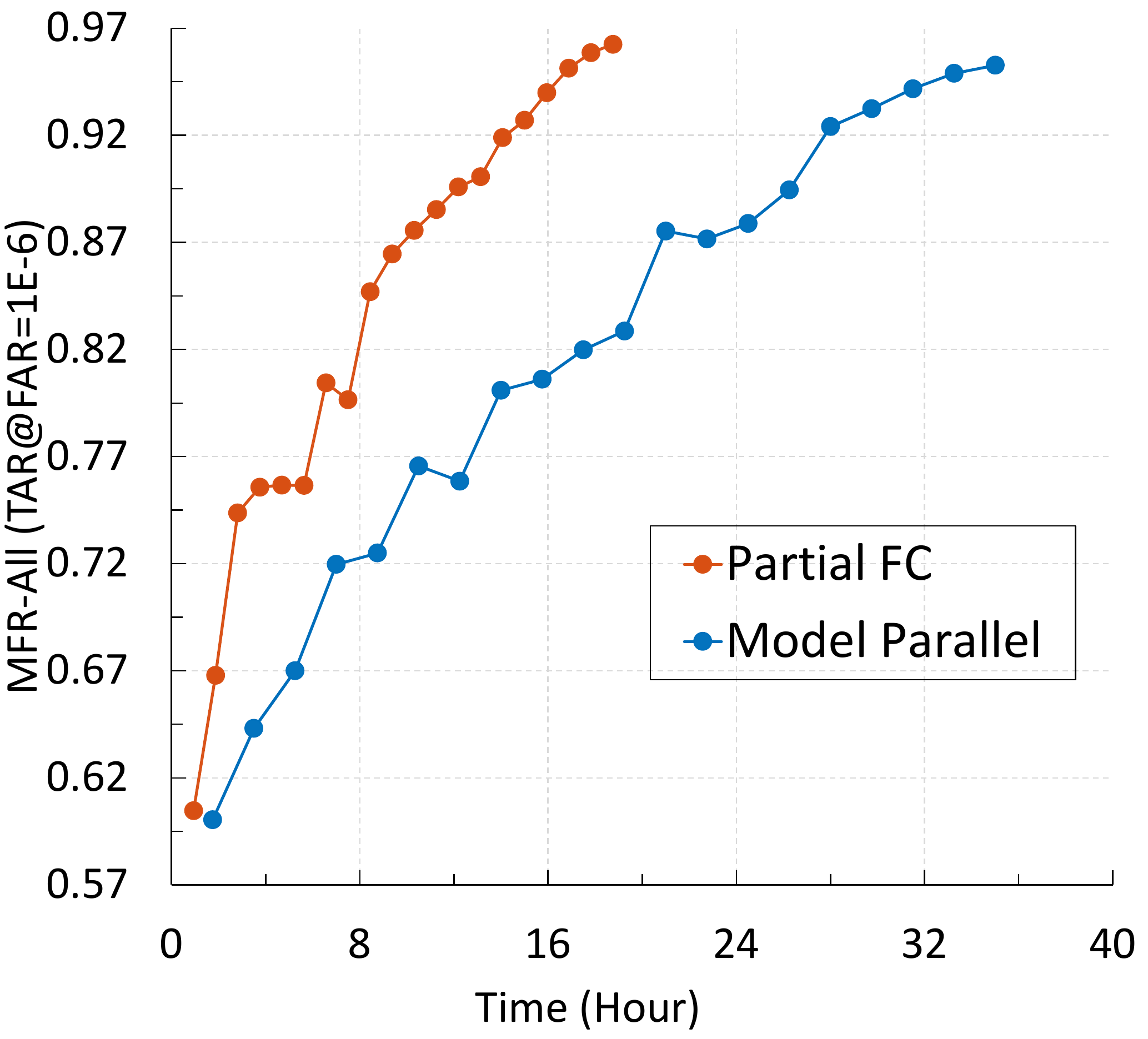}}
\vspace{-2mm}
\caption{Training status comparisons between model parallel and PFC on WebFace42M. ResNet100 is employed here. The batch size is $128\times8\times4$.}
\vspace{-4mm}
\label{fig:webface2m}
\end{figure}

\begin{table}[t]
\centering
\resizebox{1.0\linewidth}{!}{
\begin{tabular}{ l| c | c | c | c | c }
\hline
\multirow{2}{*}{Method} &
\multicolumn{3}{c| } {Verification Accuracy } &
\multicolumn{2}{c } {IJB} \cr
\cline{2-6}
                                               & LFW   & CFP-FP & AgeDB & IJB-B & IJB-C  \cr \hline 
CosFace \cite{tencent2018CosineFace}          {\small(CVPR18)} & 99.81 & 98.12  & 98.11 & 94.80 & 96.37  \cr  
ArcFace  \cite{deng2019arcface}               {\small(CVPR19)} & 99.83 & 98.27  & 98.28 & 94.25 & 96.03  \cr    
AFRN \cite{kang2019attentional}               {\small(ICCV19)} & 99.85 & 95.56  & 95.35 & 88.50 & 93.00  \cr
MV-Softmax \cite{wang2019mis}                 {\small(AAAI20)} & 99.80 & 98.28  & 97.95 & 93.60 & 95.20  \cr
GroupFace  \cite{kim2020groupface}            {\small(CVPR20)} & 99.85 & 98.63  & 98.28 & 94.93 & 96.26  \cr      
CircleLoss \cite{sun2020circle}               {\small(CVPR20)} & 99.73 & 96.02  & -     & -     & 93.95  \cr  
DUL \cite{chang2020data}                      {\small(CVPR20)} & 99.83 & 98.78  & -     & -     & 94.61  \cr
CurricularFace \cite{huang2020curricularface} {\small(CVPR20)} & 99.80 & 98.37  & 98.32 & 94.8  & 96.10  \cr
URFace \cite{shi2020towards}                  {\small(CVPR20)} & 99.78 & 98.64  & -     & -     & 96.60  \cr
DB \cite{cao2020domain}                       {\small(CVPR20)} & 99.78 & -      & 97.90 & -     & -      \cr
Sub-center \cite{deng2020sub}                 {\small(ECCV20)} & 99.80 & 98.80  & 98.31 & 94.94 & 96.28  \cr 
BroadFace \cite{kim2020broadface}             {\small(ECCV20)} & 99.85 & 98.63  & 98.38 & 94.97 & 96.38  \cr   
BioMetricNet \cite{ali2020biometricnet}       {\small(ECCV20)} & 99.80 & 99.35  & 96.12 & -     & -      \cr 
SST \cite{du2020semi}                         {\small(ECCV20)} & 99.75 & 95.10  & 97.20 & -     & -      \cr 
VPL \cite{deng2021variational}                {\small(CVPR21)} & 99.83 & 99.11  & 98.60 & 95.56 & 96.76  \cr 
VirFace \cite{li2021virface}                  {\small(CVPR21)} & 99.56 & 97.15  & -     & 88.90 & 90.54  \cr  
DCQ  \cite{bi2021dcq}                         {\small(CVPR21)} & 99.80 & 98.44  & 98.23 & -     & -      \cr  
Virtual FC \cite{li2021virtual}               {\small(CVPR21)} & 99.38 & 95.55  & -     & 67.44 & 71.47  \cr 
WebFace12M \cite{zhu2021webface260m}          {\small(CVPR21)} & 99.83 & 99.38  & 98.33 & -     & 97.51  \cr   
WebFace42M \cite{zhu2021webface260m}          {\small(CVPR21)} & 99.83 & 99.38  & 98.53 & -     & 97.76  \cr  
MC-mini-AMC \cite{zhang2021adaptive}          {\small(ICCV21)} & -     & 96.53  & 97.25 & 93.13 & 95.27  \cr  \hline
WF4M, R100, PFC-0.04                                & 99.83 & 99.06  & 97.52 & 94.91 & 96.80  \cr  
WF4M, R100, PFC-0.3                                 & 99.85 & 99.23  & 98.01 & 95.64 & 97.22  \cr 
WF12M, R100, PFC-0.013                              & 99.83 & 99.21  & 97.93 & 95.84 & 97.39  \cr 
WF12M, R100, PFC-0.3                                & 99.83 & 99.40  & 98.53 & 96.31 & 97.58  \cr  
WF42M, R100, PFC-0.008                              & 99.83 & 99.32  & 98.27 & 96.02 & 97.51  \cr 
WF42M, R100, PFC-0.3                                & 99.85 & 99.40  & 98.60 & 96.47 & 97.82  \cr 
WF42M, R200, PFC-0.3                                & 99.83 & 99.51  & 98.70 & 96.57 & 97.97  \cr 
WF42M, ViT-B, PFC-0.3                               & 99.83 & 99.40  & 98.53 & 96.56 & 97.90  \cr 
WF42M, ViT-L, PFC-0.3                               & 99.83 & 99.44  & 98.67 & 96.71 & 98.00  \cr 
\hline
\end{tabular}
}
\vspace{-2mm}
\caption{Performance comparisons between PFC and recent state-of-the-art methods on various benchmarks. 1:1 verification accuracy ($\%$) is reported on the LFW, CFP-FP and AgeDB datasets. TAR@FAR=1e-4 is reported on the IJB-B and IJB-C datasets. Under the small sampling ratios, only within-batch negative classes are used to construct the softmax loss. Therefore, the batch size of $r=0.04$ and $r=0.013$ is enlarged to $8K$, while the batch size of $r=0.008$ is amplified to $16K$.}
\label{table:allbenchmark}
\vspace{-4mm}
\end{table}

\noindent{\bf Robustness under inter-class conflict.}
In Tab.~\ref{ablationinterclassconflict}, we synthesize a WebFace12M-Conflict dataset by randomly splitting 200K identities into 600K identities, thus WebFace12M-Conflict contains 1M classes with a high inter-class conflict ratio. The performance of baseline significantly drops from $91.70\%$ to $79.93\%$ on MFR-All, while PFC is less affected by the conflicted inter-classes, demonstrating the robustness of PFC under heavy inter-class conflict.
By using an online abnormal inter-class filtering (Fig.~\ref{partial_fc} and Fig.~\ref{fig:interclassnoisesplit}), PFC*-0.2/0.3 can achieve almost the same performance as on the cleaned version of WebFace12M. Even though the baseline (FC) can also use the trick of abnormal inter-class filtering, the margin between hard inter-classes and conflicted inter-classes in FC is less clear than the proposed PFC as shown in Fig.~\ref{fig:interclassnoisesplit}. Therefore, FC* only achieves $91.18\%$ even all of the negative classes are checked at higher computation cost in each iteration.

In Tab.~\ref{ablationflipnoise}, we randomly change the image labels by adding synthesized label-flip noise into the WebFace12M dataset.
As the noise ratios increases from $0\%$ to $40\%$, the performance of baseline dramatically declines from $91.70\%$ to $43.87\%$. By contrast, the proposed PFC-0.1 is more robust under label-flip noise, achieving an accuracy of $78.53\%$ under $40\%$ label-flip noise. By using the abnormal inter-class filtering, PFC*-0.1 trained on the noisy WebFace12M ($40\%$ label-flips) further improves the verification accuracy to $80.20\%$, surpassing the FC baseline by $36.33\%$. 

\noindent{\bf Robustness under long-tail distribution.}
In Tab.~\ref{tab:longtail}, we construct a long-tail distributed dataset from WebFace42M. Specifically, 200K identities are directly copied and the rest of 1.8M identities are randomly condensed to contain 2 to 4 face images per identity. As can be seen, PFC-0.2 achieves a verification accuracy of $91.96\%$ on MFR-All, surpassing the FC baseline by $4.52\%$ and DCQ by $2.59\%$, indicating that PFC is more robust under long-tail distribution. 

\noindent{\bf Memory saving and training acceleration.}
In Tab.~\ref{tab:momandefficiency}, we compare PFC-0.1 with other sampling-based methods (\eg HF-Softmax \cite{zhang2018accelerated} and D-Softmax \cite{he2020softmax}) on WebFace42M using single computing node. Even though HF-Softmax can significantly reduce memory consumption, the time cost on feature retrieval using CPU can not be ignored. Besides, inter-class conflict still exists in the automatic cleaned WebFace42M, thus hard mining can even deteriorate the model's performance. D-Softmax separates softmax loss into intra-class and inter-class objectives and reduces the calculation redundancy of the inter-class objective. However, there exists an obvious performance drop for D-Softmax on MFR-All. In addition, the classification layer of D-Softmax uses data parallelism, thus the communication cost of the center weights will decrease the training speed. By contrast, the proposed PFC-0.1 is not only faster but also more accurate, achieving the verification accuracy of $96.19\%$.

In Tab.~\ref{tab:momandefficiency}, we also test the training speed of PFC-0.1 on synthetic 10M identities by using 8 GPUs. PFC-0.1 can run five times faster than the baseline, consuming less than half GPU memory. As logits occupy a large amount of GPU memory, the batch size of the baseline can be only $16$, resulting in a slow training speed. When the GPU number increasing from $8$ to $64$, we can observe similar phenomena of memory reduction and throughput improvement. In Fig.~\ref{fig:webface2m}, we compare the loss and performance between the FC baseline and PFC-0.1 during training. The loss of PFC in Fig.~\ref{fig:losslr} is lower than the baseline as the denominator of Eq.~\ref{eq:softmax} is smaller for PFC-0.1. In Fig.~\ref{fig:webface2mspeed}, PFC-0.1 achieves better performance than the baseline with half training time, indicating that PFC-0.1 can significantly accelerate model training on the large-scale dataset.

\subsection{Celebrity Benchmark Results}
To compare with recent state-of-the-art competitors, we train PFC models on different datasets and test on various benchmarks. As reported in Tab.~\ref{table:allbenchmark}, the proposed PFC ($r=0.3$) achieves state-of-the-art results compared to the competitors on pose-invariant verification, age-invariant verification, and mixed-media (image and video) face verification. Even though Virtual FC \cite{li2021virtual} can reduce the parameters by more than $100$ times, there is an obvious performance drop for Virtual FC on the large-scale test set. By contrast, the PFC model ($r=0.04$) trained on WebFace4M significantly outperforms the Virtual FC model by $27.47\%$ and $25.33\%$ on IJB-B and IJB-C, respectively. Both SST \cite{du2020semi} and DCQ \cite{bi2021dcq} abandon the FC layer and employ a momentum-updated network to produce class weights. However, PFC only needs to train one network instead of a pair of networks. On CFP-FP, PFC models trained on WebFace4M with sampling ratios of $r=0.04$ and $r=0.3$ outperform the DCQ model by $0.62\%$ and $0.79\%$, respectively. When WebFace42M \cite{zhu2021webface260m} is employed, the performance on all of these benchmarks tends to be saturated. However, the proposed PFC can still break the record. Specifically, the ResNet200 model trained with PFC-0.3 achieves $99.51\%$ on CFP-FP and $98.70\%$ on AgeDB, while the ViT large model trained with PFC-0.3 obtains $96.71\%$ and $98.00\%$ verification accuracy on IJB-B and IJB-C.

\section{Conclusions and Discussions}

In this paper, we propose Partial FC (PFC) for training face recognition models on large-scale datasets. In each iteration of PFC, only a small part of class centers are selected to compute the margin-based softmax loss, the probability of inter-class conflict, the frequency of passive update on tail class centers, and the computing requirement can be dramatically reduced. Through extensive experiments, we confirm the effectiveness, robustness and efficiency of the proposed PFC. 

\noindent{\bf Limitations.} Even though the PFC models trained on WebFace have achieved impressive results on high-quality test sets, it may perform poorly when face resolution is low or faces are captured in low illumination. 

\noindent{\bf Negative Societal Impact.} The PFC models may be used in surveillance and breach privacy rights, thus we will strictly control the license of code and models for academic research use only.

{\small
\bibliographystyle{ieee_fullname}
\bibliography{egbib}
}

\end{document}